\documentclass[journal]{IEEEtran}
\usepackage{graphicx}
\usepackage{amsmath}
\usepackage{amssymb,subfigure,cases}
\usepackage{setspace}
\usepackage{color}
\usepackage{hyperref}
\usepackage{algorithm}
\usepackage{algorithmicx}
\usepackage{algpseudocode}
\usepackage{multirow}
\usepackage{booktabs} 
\usepackage{url}
\usepackage{bm}
\usepackage{verbatim}
\usepackage{comment}

\usepackage[dvipsnames]{xcolor}

\newcommand{\R}{\mathbb{R}}

\ifCLASSOPTIONcompsoc
  \usepackage[nocompress]{cite}
\else
  \usepackage{cite}
\fi

\ifCLASSINFOpdf
\else
 \fi

\begin{document}
%
% paper title
% Titles are generally capitalized except for words such as a, an, and, as,
% at, but, by, for, in, nor, of, on, or, the, to and up, which are usually
% not capitalized unless they are the first or last word of the title.
% Linebreaks \\ can be used within to get better formatting as desired.
% Do not put math or special symbols in the title.
\title{THE Benchmark: Transferable Representation Learning for Monocular Height Estimation}
\author{
     Zhitong~Xiong, \IEEEmembership{Member,~IEEE}, Wei Huang, Jingtao Hu, and Xiao Xiang Zhu, \IEEEmembership{Fellow,~IEEE}
\IEEEcompsocitemizethanks{
\IEEEcompsocthanksitem This work is jointly supported by German Federal Ministry for Economic Affairs and Climate Action in the framework of the "national center of excellence ML4Earth" (grant number: 50EE2201C), by the German Research Foundation (DFG GZ: ZH 498/18-1; Project number: 519016653), by the German Federal Ministry for the Environment, Nature Conservation, Nuclear Safety and Consumer Protection (BMUV) based on a resolution of the German Bundestag (grant number: 67KI32002B; Acronym: \textit{EKAPEx}), and by the German Federal Ministry of Education and Research (BMBF) in the framework of the international future AI lab "AI4EO -- Artificial Intelligence for Earth Observation: Reasoning, Uncertainties, Ethics and Beyond" (grant number: 01DD20001). (Corresponding author: Xiao Xiang Zhu.)
\IEEEcompsocthanksitem Z. Xiong, W. Huang, and X. X. Zhu are with the chair of Data Science in Earth Observation, Technical University of Munich (TUM), 80333 Munich, Germany. (e-mails: zhitong.xiong@tum.de; w2wei.huang@tum.de;  xiaoxiang.zhu@tum.de)
\IEEEcompsocthanksitem J. Hu is with the School of Artificial Intelligence, Optics and Electronics (iOPEN), Northwestern Polytechnical University, 710072 Xi'an, China (e-mails: jthu@mail.nwpu.edu.cn)
}}
\markboth{IEEE TRANSACTIONS ON Geoscience and Remote Sensing,~Vol.~XXX, No.~XXX, XXX~XXX}%
{Shell \MakeLowercase{\textit{et al.}}: Bare Advanced Demo of IEEEtran.cls for IEEE Computer Society Journals}
\maketitle

\begin{abstract}
Generating 3D city models rapidly is crucial for many applications. Monocular height estimation is one of the most efficient and timely ways to obtain large-scale geometric information. However, existing works focus primarily on training and testing models using unbiased datasets, which does not align well with real-world applications. Therefore, we propose a new benchmark dataset to study the transferability of height estimation models in a cross-dataset setting. To this end, we first design and construct a large-scale benchmark dataset for cross-dataset transfer learning on the height estimation task. This benchmark dataset includes a newly proposed large-scale synthetic dataset, a newly collected real-world dataset, and four existing datasets from different cities. Next, a new experimental protocol, \emph{few-shot cross-dataset transfer}, is designed. Furthermore, in this paper, we propose a scale-deformable convolution module to enhance the window-based Transformer for handling the scale-variation problem in the height estimation task. Experimental results have demonstrated the effectiveness of the proposed methods in traditional and cross-dataset transfer settings. The datasets and codes are publicly available at \url{https://mediatum.ub.tum.de/1662763} and \url{https://thebenchmarkh.github.io/}.
\end{abstract}

% Note that keywords are not normally used for peerreview papers.
\begin{IEEEkeywords}
Cross-dataset transfer, remote sensing, synthetic data, transfer learning, transformer, benchmark
\end{IEEEkeywords}

% make the title area

\section{Introduction}
\label{introduction}
\IEEEPARstart{M}{onocular} height estimation (MHE) \cite{mou2018im2height} is of great importance to rapid 3D city modeling and can give a basic insight into urbanization level. Geometric information from 3D cities can be used for energy demand estimation, population estimation, damage forecasting and so on \cite{navalgund2007remote}. On the other hand, the generated height data can also contribute to many challenging follow-up research topics, such as urban planning \cite{masser2001managing}, automatic piloting, and robot vision.

Airborne Light Detection And Ranging (LiDAR) can actively acquire the Digital Surface Model (DSM) data that contains accurate height information. However, LiDAR is cost-consuming, due to the airborne carrying platform and data storage device. Furthermore, its height information is not updated in a timely manner,  limited by the hardware and complicated post-processing process. Apart from LiDAR, incidental satellite images can also provide height data via the calculation of triangulation from satellite image pairs of consecutive views at different time intervals. Nevertheless, the conditions for obtaining such images are quite strict. In addition, both LiDAR and incidental images are difficult to handle complex city scenes in a real-time manner. In contrast, MHE can predict height using only a single aerial image, and offers broad application potential in practice because of its fairly simple data acquisition requirements. %However, there are still some issues that hinder the development of this research. 

\begin{figure}
	\begin{center}
		\includegraphics[width=0.48\textwidth]{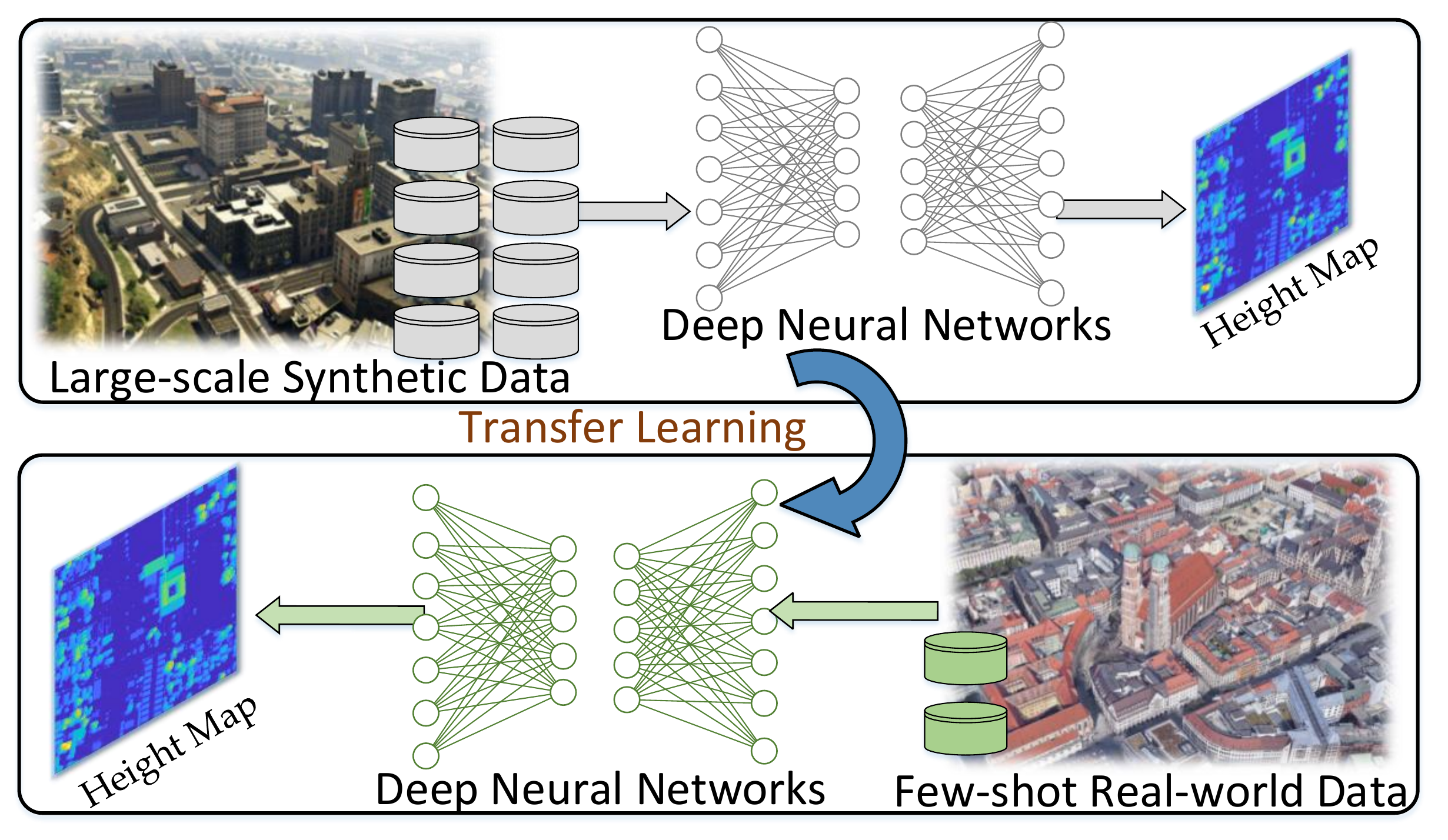}
	\end{center}
	\caption{Illustration of monocular height estimation in the cross-dataset transfer setting. We aim to transfer deep models from a large-scale synthetic dataset to different real-world datasets in a few-shot cross-dataset setting.}
	\label{motivation1}
\end{figure}
Benefiting from the powerful feature representation capacity, deep neural networks (DNN) \cite{lecun1998gradient} have dominated most computer vision fields, including MHE. However, deep learning-based MHE methods usually require a large amount of annotated data, which is difficult to obtain for several reasons.

Firstly, it is very costly to obtain high-resolution remote sensing images and their corresponding ground truth height values on a global scale. Due to different development levels and urban construction styles, different cities have their specific urban layouts, which leads to distinctive height distribution. This results in severe cross-city domain shifts that are very common in real-world applications. One potential approach to mitigate these domain shifts is to gather data from multiple cities. However, it is prohibitively expensive or even impossible to collect high-quality data samples for the MHE task from diverse cities at a global scale.

Secondly, to ensure the performance of MHE models in real-world applications, it is imperative to construct datasets for MHE model training that encompass a wide range of imaging conditions. Existing methods usually train and test MHE models using unbiased datasets \cite{yosinski2014transferable}. However, for real-world applications, providing the training data under diverse imaging conditions can be effective at improving the robustness and performance of deep networks. However, constructing datasets with different imaging conditions, like different camera poses (heights and angles), camera resolutions, and viewing fields, is further expensive and difficult. Consequently, there is a lack of a high-resolution, highly-accurate, and large-scale annotated height estimation benchmark dataset.

%Given this problem, in this work, we take the city biases into consideration by proposing a new cross-city transfer benchmark dataset. We also benchmark some baseline results to provide valuable insights on designing height estimation models in a cross-dataset setting. 

To address these aforementioned limitations, we resort to constructing a large-scale synthetic dataset that contains high-resolution images with accurate geometric information captured under different conditions. The presented benchmark dataset, termed THE, can foster research on transferable representation learning for monocular height estimation. In addition to the benchmark dataset, we design a new Transformer-based method to enhance the performance of MHE models in two cross-dataset experimental settings. To summarize, we make the following contributions:

\begin{itemize}
\item [(1)]  Collecting and releasing two new datasets for MHE. One is a large-scale synthetic dataset termed GTAH (Grand Theft Auto for Height estimation), which is obtained from the game \emph{Grand Theft Auto} \cite{GTAV}, under different imaging conditions. %GTAH contains 28,627 height maps in total and each with a resolution of 1920$\times$1080. For each height map, there are three corresponding RGB images that are captured under different weather conditions. Besides the weather, there are many other imaging conditions that have been taken into consideration, such as the daytime, shadow, camera height, and pose. 
The other dataset is a real-world one collected from the Actueel Hoogtebestand Nederland (AHN) project, which covers multiple cities in the Netherlands.
%In the AHN dataset, 10,755 images with a resolution of 1024$\times$1024 are carefully selected and processed. 

\item [(2)] Constructing a new benchmark platform for transferable monocular height estimation. Specifically, one synthetic dataset and five real-world datasets are included to explore the feasibility of height knowledge transfer from synthetic to real scenes. We propose a few-shot cross-dataset transfer setting to evaluate deep models on datasets that were not seen during training.
%, as shown in Fig. \ref{motivation1}.

\item [(3)] To further enhance the model transferability in a cross-dataset experimental setting, we design a new scale-deformable convolution (SDC) module to enhance the Transformer networks with adaptive spatial context. The SDC module can learn to adjust the spatial context of representations adaptively across different datasets.
\end{itemize}

Extensive quantitative and qualitative results show that our framework outperforms existing methods clearly, which indicates the effectiveness of the proposed methods. The remainder of this paper is organized as follows. Section~\ref{related work} reviews related works. Section~\ref{methodology} introduces the details of the proposed method. In Section~\ref{experiments}, extensive experiments and analysis are presented to verify the proposed method comprehensively. Finally, this work is concluded in Section~\ref{conclusion}.

\section{Related Work}
\label{related work}
Both monocular depth estimation (MDE) and MHE are geometry-related regression tasks; the former motivates the development of the latter to some extent. In this section, related works on MDE are firstly investigated and then MHE is introduced.

\textbf{Monocular Depth Estimation}. Early works on MDE utilized hand-crafted visual features and probabilistic graphical models (PGMs) to encode depth-specific visual cues, including object size and texture density, based on a strong geometric assumption \cite{saxena2005learning, saxena2008make3d, konrad20122d}. Recently, deep learning-based methods have dominated this field because of their powerful feature representation capacity. There are roughly two types of deep learning-based MDE, supervised methods \cite{eigen2015predicting,zhu2020edge,jiao2018look,miangoleh2021boosting} and self-supervised (unsupervised) methods \cite{garg2016unsupervised, godard2017unsupervised, pilzer2019refine, peng2021excavating, jung2021fine, jung2021fine, jiao2021effiscene, tosi2020distilled, godard2019digging, shu2020feature, guizilini20203d, lyu2021hr, guizilini2019semantically}. 

In this work, we mainly introduce the supervised methods, which take a single image as input and generate a pixel-wise depth prediction map, following the standard supervised-learning workflow with the need for manual-annotated depth labels. These methods achieve state-of-the-art performance by making breakthroughs in innovative architecture designs, effective incorporation of geometric and semantic constraints, and novel objective functions. Some enlightening works \cite{liu2015learning, laina2016deeper, eigen2015predicting} have applied deep convolutional neural network (CNN) architectures to MDE, directly estimating depths from single monocular images in an end-to-end trainable manner, and achieving impressive performance. To model the semantic and geometric structure of objects within a scene, some work \cite{chen2019towards, wang2020sdc, yin2019enforcing} has introduced semantic segmentation into MDEs as an auxiliary task, which can guide depth estimation at the object level. 

Taking into account the imbalanced depth distribution that restricts model performance, Jiao et al. \cite{jiao2018look} presented an attention-based distance-related loss to concern more distant depth regions. Lee et al. \cite{lee2020multi} combined multiple loss terms adaptively to train a monocular depth estimator from a constructed loss function space containing many loss terms. To balance the coverage speed of these losses, a loss-aware adaptive rebalancing algorithm was further designed in the course of training. The work most closely related to ours is \cite{ranftl2019towards}, in which a robust training objective is designed to train deep learning-based MDE models using multiple mixing datasets. For the first time, they propose to evaluate MDE models in a zero-shot cross-dataset transfer setting. More recently, vision Transformer-based deep models \cite{ranftl2021vision, li2021revisiting} have also been proposed, to take advantage of the powerful representation learning ability of the Transformer backbone.

%To gain a deeper understanding of exactly what visual cues really influence the MDE performance, Dijk et al. \cite{dijk2019neural} made extensive comparison experiments on multiple published networks to study the roles of several visual cues when estimating depths, including the position and apparent size of objects, camera poses, and obstacle interference.

%On the other hand, researchers have gradually paid more attention to self-supervised methods in order to eliminate the heavy dependence on large-scale depth labels that are cumbersome and formidable. This type of methods avoid the demand for label data by converting depth estimation to a reconstruction task, where the depth map is an intermediate among stereo images, sequential videos or their combination. 
%For stereo images, the deep model with their known relative placements only needs to predict their disparity map, i.e. the inverse of the depth map \cite{garg2016unsupervised, godard2017unsupervised, pilzer2019refine, peng2021excavating, jung2021fine}. 
%For sequential videos, additional predictions of the relative pose of the camera are required \cite{zhou2017unsupervised, yin2018geonet, jung2021fine, jiao2021effiscene, tosi2020distilled}. 
%Similar to improvement strategies of supervised MDE, related works on self-supervised MDE have dramatically boosted the prediction performance, centering on new loss function \cite{godard2019digging, shu2020feature}, novel architecture \cite{guizilini20203d, lyu2021hr}, and extra supervision from other constraints \cite{guizilini2019semantically, poggi2020uncertainty}.
\begin{figure*}[!]
	\begin{center}
		\includegraphics[width=0.935\textwidth]{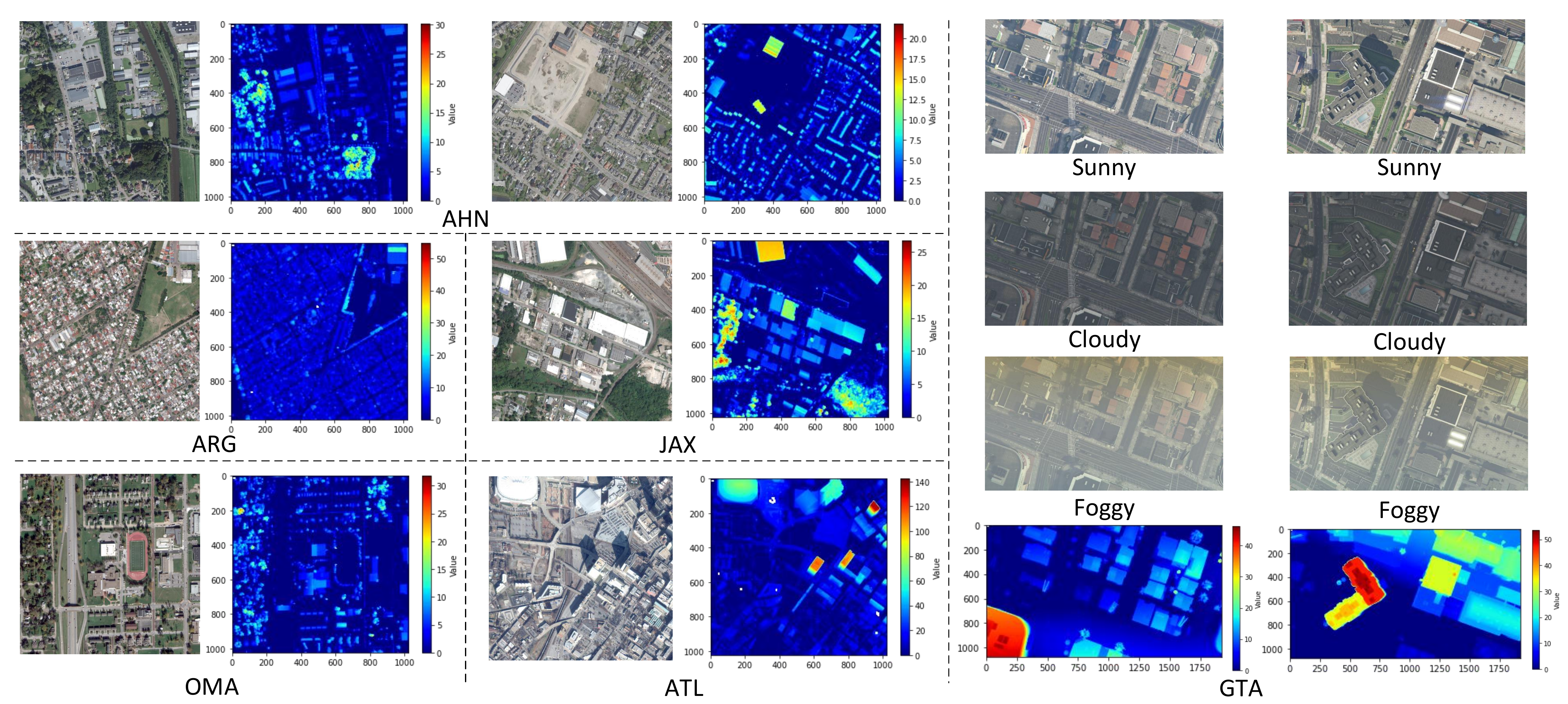}
	\end{center}
	\caption{Visualization of some samples from the synthetic GTAH dataset and real-world datasets. To simulate the complex real-world conditions in monocular aerial images, various imaging conditions, which may influence the performance of MHE, are taken into consideration, including different heights, weathers, and times of day. Moreover, it can be seen that the scale-variation problem between different datasets is very obvious.}
	\label{ahn_show}
\end{figure*}
\textbf{Monocular Height Estimation}. 
Motivated by the success of deep learning-based MDE, researchers have attempted to directly predict the height of objects, i.e., the digital surface model (DSM), within single aerial images from an overhead view \cite{srivastava2017joint, mou2018im2height, ghamisi2018img2dsm, paoletti2020u, amini2019cnn, amirkolaee2019height, liu2021hecr, christie2020learning, mahmud2020boundary, mahmud2020boundary, madhuanand2021self, zheng2019pop, liu2020im2elevation, li2020height}. 
Srivastava et al. \cite{srivastava2017joint} proposed a multi-task CNN architecture for joint height estimation and semantic segmentation in monocular aerial images in 2017 for the first time. 
Mou and Zhu \cite{mou2018im2height} published a concurrent work that proposed a fully convolutional-deconvolutional network for MHE and demonstrated its usefulness, for instance, the segmentation of buildings. Paoletti et al. \cite{ghamisi2018img2dsm, paoletti2020u} performed the image-to-image translation from monocular optical images to the corresponding depth maps within three cities, using the technique of generative adversarial network (GAN). Amini and Arefi \cite{amini2019cnn} presented a CNN-based method to identify collapsed buildings after an earthquake, based on pre-event and post-event satellite images as well as airborne LiDAR data. Based on the CNN architecture, they further designed a post-processing approach to merge multiple predicted height image patches into a seamless continuous height map \cite{amirkolaee2019height}. Liu et al. \cite{liu2021hecr} proposed a joint framework called height-embedding context reassembly network (HECR-Net) to simultaneously predict semantic labels and height maps from single aerial images, by distilling height-aware embeddings implicitly.

Leveraging the optical flow prediction technique, Chridtie et al. \cite{christie2020learning} developed an encoding strategy of the universal geocentric pose of objects within static monocular aerial images, and trained a deep network to compute the dense representation; these attributes were exploited to rectify oblique images to dramatically improve the accuracy of height prediction of multiple images taken from different oblique viewpoints.
Mahmud et al. \cite{mahmud2020boundary} proposed a boundary-aware multi-task deep-learning-based architecture for fast 3D building modeling from single overhead images, by jointly learning a modified signed distance function, a dense height map, and scene semantics from building boundaries in order to model the buildings within the scenes.
Madhuanand et al. \cite{madhuanand2021self} aimed to estimate depth from a single Unmanned Aerial Vehicle (UAV) aerial image, by designing a self-supervised learning approach named Self-supervised Monocular Depth Estimation that does not need any information other than images.

Although these works have contributed to the development of MHE in the past few years, most of them stayed within a particular comfort zone. There is an urgent need to study some more crucial issues that restrict the practical application of MHE in the open real world, such as the exploration in few-shot knowledge transfer in a cross-dataset setting, and scale-adaptive MHE model design. Aiming to address these problems, this paper conducts the corresponding research and exploration.

\section{Transferable Monocular Height Estimation}
This section describes the proposed \textbf{T}ransferable Monocular \textbf{H}eight \textbf{E}stimation (THE) benchmark, which includes a synthetic GTAH dataset and five real-world datasets. The newly constructed GTAH and AHN datasets are described in detail, including their data source, dataset details, and statistical characteristics. Then to fairly evaluate the proposed two data sets in the MHE field, a comprehensive statistical analysis for datasets from five different domains is performed.

\subsection{GTAH}
In this subsection, the data source and dataset details of the synthetic GTAH data set are introduced in detail.

\textbf{Data Source: } GTAH was collected from an electronic computer game called Grand Theft Auto V (GTA5) that was developed by Rockstar North and published by Rockstar Games \cite{GTAV}. The virtual world in GTA5 covers an area of 252 square kilometers, containing many scenes like beach, stadium, mall, store, and so on.

\textbf{Dataset Details: } GTAH contains a total of 85,881 pairs of synthetic monocular aerial images and their associated pixel-wise height maps, with a resolution of 1920$\times$1080. To simulate the complex real-world conditions in monocular aerial images, various imaging conditions are taken into consideration, including:

\begin{itemize}
	\item [(1)] 
	 \textbf{Height distribution}. Most of the scenes in GTAH are located in areas with a rich variety of buildings for height diversity. Diverse height  facilitates a comprehensive and fair assessment of the performance of MHE algorithms, for the situation where height estimation is valid for some heights but is poor when faced with a wide variety of heights. 

	\item [(2)] 
	\textbf{Camera locations}. For the diversity of height and scene information, 1,111 positions are selected along the roads in GTA5's city as the plane coordinates of the camera without respect to camera heights. 
	
	\item [(3)]
	\textbf{Camera angles}. Taking [\textit{x}, \textit{y}, \textit{z}] as the coordinate system, monocular aerial images of GTAH are captured from different viewpoints in order to simulate the diversity of camera pose.

	\item [(4)]
	\textbf{Camera heights}. For the diversity of scene scales, monocular images are acquired at four camera heights of 300, 380, 460, and 540 meters, to study the effects of the camera height in practice. 
	
	\item [(5)]
	\textbf{Weather types}. To evaluate the effectiveness and robustness of MHE methods in different weathers, GTAH contains three common weather conditions: ``sunny," ``foggy, " and ``cloudy." 
	
	\item [(6)]
	\textbf{Shadows}. Shadow is an implicit visual cue influencing the performance of MDE models. In \cite{dijk2019neural}, Dijk et al. made an ablation study of shadows to demonstrate their effect on MDE. As a similar task focusing on pixel-wise regression, it could be presumed that MHE may also be influenced by shadows. To enable a study of this kind, GTAH contains the monocular images with (w/) and without (w/o) shadows.

	\item [(7)]
	\textbf{Capturing different times of day}. Different times of day have a significant impact on light intensity and direction, which further determines the shadow direction of buildings. For the fine study of capturing times of day and their subsequent effects, three times of day are considered in GTAH: 9:00 AM, 15:00 PM, and 18:00 PM.
	
\end{itemize}
Some statistical results of GTAH are shown in Fig. \ref{spatial_compare} and some examples of GTAH are illustrated in Fig. \ref{ahn_show}.

\begin{table*}[]
	\centering
	\label{CompData}
	\caption{Comparison between the proposed datasets and other existing MHE datasets}
	\begin{tabular}{c|c|c||c|c|c|c}
		\hline
		{Datasets}         & \textbf{GTAH (Ours)} & \textbf{AHN (Ours)} & JAX (US3D)       & OMA (US3D)       & ATL (US3D)       & ARG (OGC)       \\ \hline
		{Number of Images} & \textbf{85,881}   & \textbf{10,755}      & 1,098      &1,796      & 704       & 2,325      \\ \hline
		{Image Size}       & 1920$\times$1080   & 1024 $\times$ 1024  & 1024 $\times$ 1024 & 1024$\times$1024 & 1024$\times$1024 & 1024$\times$ 1024 \\ \hline
		{Vertical/Oblique} & {Vertical \& Oblique}    & Vertical   & Oblique   & Oblique   & Oblique   & Oblique   \\ \hline
		{Weather Types}    & \textbf{Multiple}    & Single     & Single    & Single    & Single    & Single    \\ \hline
		{Daytimes}         & \textbf{Multiple}    & Single     & Single    & Single    & Single    & Single    \\ \hline
        {{Height Scale}}         & {[0,439.2]}    & {[0,195.8]}     & {[0,186.5]}    & {[0,194.1]}    & {[0,123.4]}    & {[0,92.7]}    \\ \hline
		{Shadows}          &\textbf{w/ \& w/o}   & w/         & w/        & w/        & w/        & w/        \\ \hline
		{Real/Synthetic}    & Synthetic   & Real       & Real      & Real      & Real      & Real      \\ \hline
		{Number of Cities}    & Single   & \textbf{Multiple}       & Single      & Single      & Single      & Single      \\ \hline
	\end{tabular}
\end{table*}

\subsection{AHN}
The data source, collection, and properties of the real-world AHN data set are introduced in this subsection.

\textbf{Data Source: } AHN was collected from the Actueel Hoogtebestand Nederland (AHN) \footnote{\url{https://www.ahn.nl/het-verhaal-van-ahn}} project.

\textbf{Dataset details: } AHN contains a total of 10,775 pairs of real monocular aerial images and their associated pixel-wise height maps, with a resolution of 1024$\times$1024. In the AHN dataset, images are selected to cover different scene types, including buildings, farms, forests, and water areas. The corresponding height maps are also carefully processed for MHE model training and evaluation. In addition, unlike other datasets, the AHN dataset covers multiple cities in the Netherlands, as shown in Fig. \ref{ahn_show}. 

\textbf{How the height maps are generated:} GTAH is generated using the GTA game, where we manipulate the pose of cameras and adjust other rendering parameters to obtain a variety of RGB images along with their corresponding height maps. The height data in GTAH is synthetically generated using a game engine, ensuring high quality and precision. The AHN dataset is acquired from the AHN project, which utilizes airborne LiDAR technology. The height data in the AHN project undergoes extensive quality control checks before being released, ensuring its reliability and accuracy. For the US3D datasets, the height data is derived from airborne LiDAR data obtained from the Homeland Security Infrastructure Program \footnote{https://www.ahn.nl/hoe-werkt-het-inwinnen-van-hoogtegegevens}. Specifically, the above ground level (AGL) height images are considered as the ground truth height data.

\subsection{Comparison with Other Existing MHE Datasets}

\begin{figure}
	\begin{center}
		\includegraphics[width=0.485\textwidth]{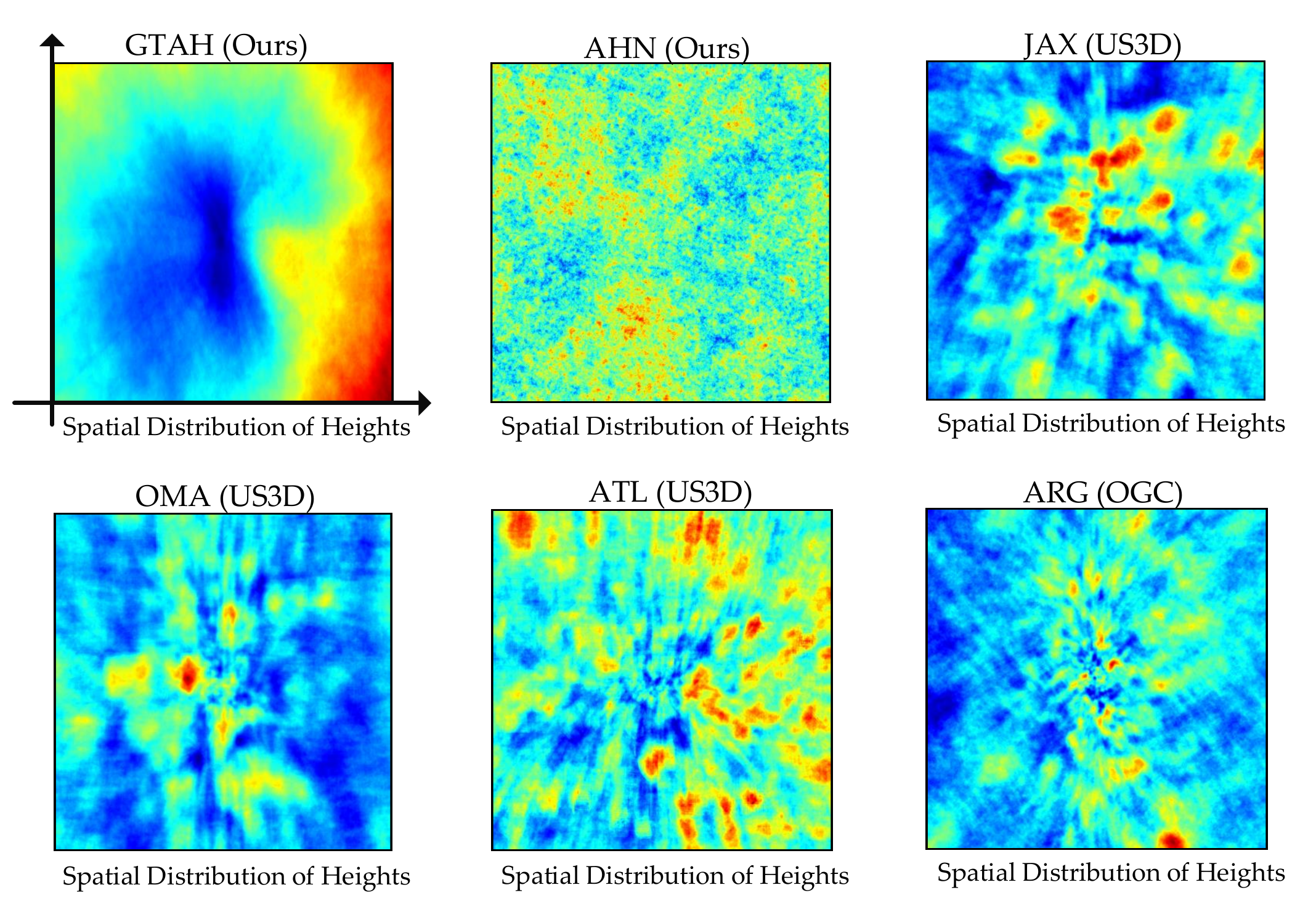}
	\end{center}
	\caption{Spatial distributions of heights of MHE data sets, including GTAH, AHN, JAX, OMA, ATL, and ARG. The x and y axes represent the width and height of the image respectively. Darker colors represent larger height values. Different cities have clearly different spatial patterns.}
	\label{spatial_compare}
\end{figure}

\begin{figure}
	\begin{center}
		\includegraphics[width=0.485\textwidth]{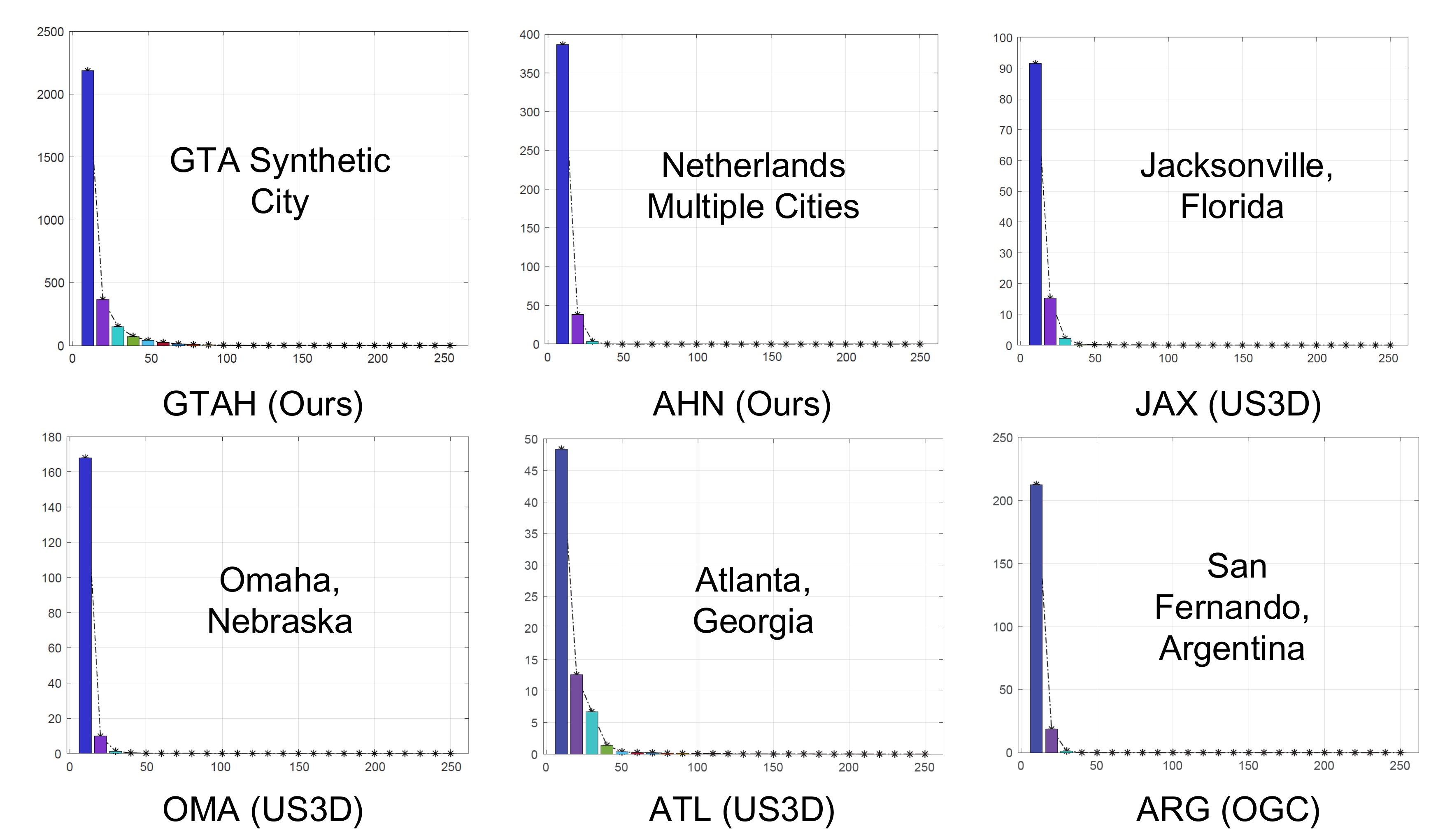}
	\end{center}
	\caption{Height distributions of six different MHE datasets, including GTAH, AHN, JAX, OMA, ATL, and ARG.}
	\label{dist_compare}
\end{figure}

To enrich the proposed THE benchmark dataset, we further take in the data of four cities from Urban Semantic 3D \cite{christie2021geocentricpose,bosch2019semantic,kunwar2020large}, including Jacksonville (JAX), Omaha (OMA), Atlanta (ATL), and Argentina (ARG). The detailed comparisons of these six MHE datasets are presented in Table \ref{CompData}.
In Fig. \ref{spatial_compare}, we visualize the spatial distributions of the height in different datasets. The ARG dataset is collected from the Overhead Geopose Challenge (OGC). \footnote{\url{https://www.nasa.gov/overhead-geopose-challenge}}
Additionally, we have also analyzed the histogram of height distribution in six different datasets. In Fig. \ref{dist_compare} we can see that the height distributions of all six datasets obey long-tail distribution.  It can be seen that the differences among these six datasets are clearly apparent. This also clearly indicates the domain shifts between different cities and the difficulty of cross-dataset transfer setting. 

\section{Methodology}
\label{methodology}
\begin{figure*}
	\label{pipeline}
	\begin{center}
		\includegraphics[width=0.95\textwidth]{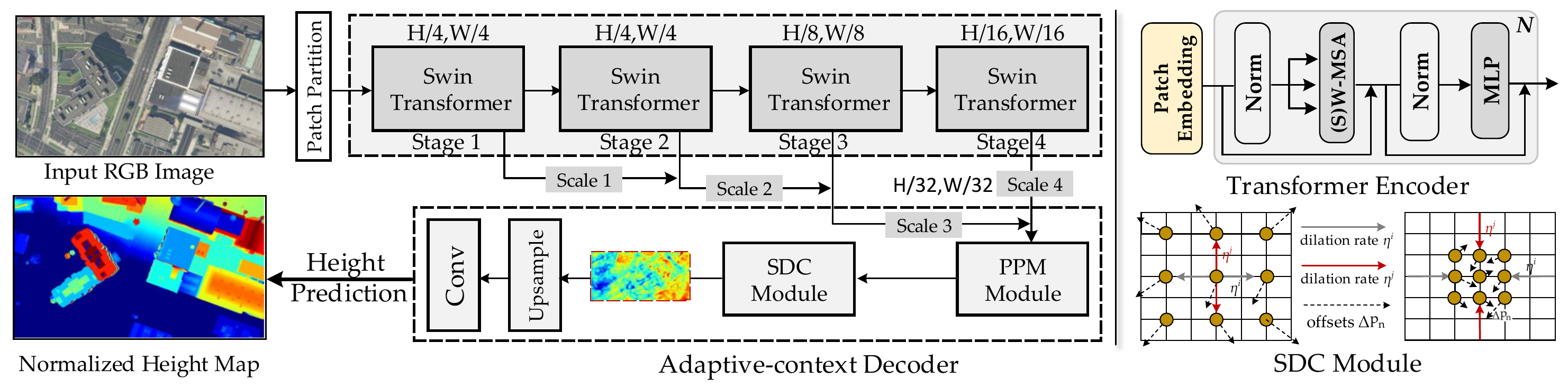}
		\caption{The whole pipeline of the proposed ``SwinUper+SDC." First, the height-related visual semantic feature is extracted by Swin Transformer. Then, adaptive scale modeling for covering a dynamic receptive field is achieved by the SDC module. Finally, the model is trained for height estimation using different types of loss functions for both the classical and cross-shot transfer settings.}
	\end{center}
\end{figure*}

In this section, we will introduce the proposed transformer-based frameworks for monocular height estimation. The whole network architecture is shown in Fig. \ref{pipeline}. Specifically, we will first introduce some existing vision transformers and their limitations for the MHE task. Then, an adaptive-structure convolution module is introduced to improve the performance and transferability of MHE. Finally, based on the constructed synthetic dataset, a few-shot cross-dataset transfer learning method is designed.

\subsection{Vision Transformer for Height Prediction}
{Compared with CNN-based deep architectures, Transformers enable relationship modeling between input tokens and can capture relative height information through the self-attention mechanism. To predict the pixel-wise height value from monocular images accurately, it is beneficial to make use of the relative height relationships between neighboring pixels. For the MHE task, this advantage makes Transformer-based architectures more effective in improving both the performance and the transferability of deep models.

\subsection{Scale-deformable Convolution for Few-shot Cross-dataset Transfer}
For real-world applications, few-shot cross-dataset performance is a more faithful evaluation metric than training and testing on datasets with the same biases. Compared with the monocular depth estimation task, the cross-dataset evaluation for MHE is more challenging. The reason is that remote sensing imagery captured at different heights will be greatly different due to the change in resolutions. However, the height values of objects on the ground should not change with different camera poses. This inconsistency makes MHE an extremely challenging task, especially in cross-dataset evaluation settings.

The Swin Transformer can greatly reduce computational complexity by computing self-attention maps within local windows. The window size is an important hyperparameter for the window-based Transformer models. However, for images with significantly different resolutions, aggregating context information in fixed-size windows has an obvious limitation: the spatial context for objects with different scales will be inconsistent. This makes the standard window-based Transformer less effective for handling the scale-variation problem. Consequently, the severe scale-inconsistent problem in remotely sensed images (as shown in Fig. \ref{ahn_show}) makes the fixed window size for image partitioning ineffective. 

Considering this limitation, in this work, we propose a scale-deformable convolution (SDC) module to adjust the spatial context of each pixel for the Transformer model in a learnable way. We achieve this goal by designing a deformable convolution operation with learnable dilation rates to adjust the context in a structured way. Given the input feature map $\textbf{X} \in \R^{c\times h\times w}$ and kernel weight $\bm{w} \in \R^{c_{\text{o}}\times c\times k\times k}$, the standard convolution can be formulated as:
\begin{equation}
{\bm{V}_{\bm{p}_0}} = \sum\limits_{\textbf{p}_n \in {\Omega}} {\bm{w}({\bm{p}_n})\cdot{\bm{x}(\bm{p})}},
\end{equation}
where $\bm{V}_{\bm{p}_0} \in \R^{c}$ denotes the output features at pixel $\bm{p}_0$.
Indexes of the 2D spatial offsets for the convolution operation are denoted by $\Omega$. For a point $\bm{p}_0$ in the output feature map, the coordinates used for convolution computation are $\textbf{p}=\textbf{p}_0+\bm{p}_n$.

To adjust the spatial context, the deformable convolution \cite{dai2017deformable} was proposed to learn additional offsets in order to get a more flexible receptive field for each pixel. Although deformable convolution can learn adaptive context by offsets, we argue that merely using the offsets is still inefficient to adjust the receptive field with significant scale variation. For the traditional deformable convolution, usually, multiple deformable convolution layers are required to gradually adjust the context by the learned offsets. While using a learnable multiplier for the receptive field would be more effective, especially in the few-shot transfer settings. Thus, in this work, we further extend the deformable convolution with learnable dilation rates. Based on this idea, the coordinates of point $p$ becomes 
\begin{equation}
\begin{split}
    \bm{p^i} = \bm{p}_0^i+\eta^i\bm{p}_n^i+\Delta\bm{p}_n^i,\\
    \bm{p^j} = \bm{p}_0^j+\eta^j\bm{p}_n^j+\Delta\bm{p}_n^j,
\end{split}
\end{equation}
where $\eta=\{\eta^i,\eta^j\}$ are the learnable dilation rates, which can be used to control the receptive field for each pixel in a structured manner. The 2D offsets of the deformable convolution are expressed by $\Delta \bm{p}_n=\{\Delta \bm{p}_n^i,\Delta \bm{p}_n^j\}$.

In practice, the dilation rates $\eta$ and offsets $\Delta \bm{p}_n$ are typically fractional. To enable their end-to-end optimization, we adopt differentiable bilinear sampling to perform scale-deformable convolution, which can be defined as   
\begin{equation}
\begin{split}
\bm{V}_{\bm{p}_0}^c = \sum\limits_u^H \sum\limits_v^W \bm{x}^c(u,v)\, &\text{max}(0,1 - |{\bm{p}^i} - v|) \\
&\text{max}(0,1 - |\bm{p}^j - u|), 
\end{split}
\label{dbi}
\end{equation}
where ${\bm{p}_0 \in \{1,2,...,HW\}}$ is the index of output feature maps, and ${c \in [1...C]}$ is the index of feature channels. For the sake of simplicity, the coordinate $\bm{p}_0$ will be omitted in the following formulas. The coordinates of input feature maps are denoted by $u,v$. Note that, coordinates ${(\bm{p}^i,\bm{p}^j)}$ and ${(u,v)}$ are normalized in the range of [-1,1].
During backward propagation, we need to compute the partial derivatives w.r.t. $\bm{x}^c(u,v)$, $\bm{p}^i$, $\bm{p}^j$, $\Delta \bm{p}_n^i$, $\bm{p}_n^j$, and $\eta^i$, $\eta^j$. Based on Eq. \ref{dbi}, the partial derivatives for $\bm{x}^c(u,v)$ can be easily obtained by   
\begin{equation}
\begin{split}
\frac{{\partial \bm{V}^c}}{{\partial \bm{x}^c(u,v)}} = \sum\limits_u^H \sum\limits_v^W &\text{max}(0,1 - |{\bm{p}^i} - v|) \\ 
&\text{max}(0,1 - |{\bm{p}^j} - u|).
\end{split}
\end{equation}
Next, the partial derivatives of $\bm{p}^i$ can be computed by 
\begin{equation}
\label{e7}
\frac{{\partial \bm{V}^c}}{{\partial {\bm{p}^i}}} = \sum\limits_{u,v}^{H,W} \bm{x}^c(u,v)\, {\text{max}(0,1 - |{\bm{p}^j} - u|)\, g(v,{\bm{p}^i})} ,
\end{equation}
where $g(v,\bm{p}^j)$ is a piecewise function that can be formulated as
\begin{equation}
\label{e8}
g(v,{\bm{p}^i}) =
\begin{cases}
0, & \text { if }\left|v-\bm{p}_{i}\right| \geq 1 \\
1, & \text { if } v \geq \bm{p}_{i} \\
-1, & \text { if } v<\bm{p}_{i}
\end{cases}
\end{equation}

The partial derivative $\frac{{\partial \bm{V}_{k}^c}}{{\partial {\bm{p}^j}}}$ is similar to that of $\bm{p}^i$. Furthermore, the partial derivative of $\eta^i$ can be obtained by applying the chain rule 
\begin{equation}
\label{e7}
\begin{split}
\frac{{\partial \bm{V}^c}}{{\partial {\eta^i}}} &= \frac{{\partial \bm{V}^c}}{{\partial {\bm{p}^i}}} \frac{{\partial \bm{p}^i}}{{\partial {\eta^i}}},\\
\frac{{\partial \bm{p}^i}}{{\partial {\eta^i}}} &= \bm{p}_n^i.
\end{split}
\end{equation}
The computation of partial derivatives $\frac{{\partial \bm{V}^c}}{{\partial {\eta^j}}}$ is similar to that of $\frac{{\partial \bm{V}^c}}{{\partial {\eta^i}}}$. Finally, we follow the same formulas described in DCN \cite{dai2017deformable} to compute the partial derivatives $\frac{{\partial \bm{V}^c}}{{\partial {\Delta \bm{p}_n}}}$.

\subsection{Scale-invariant Training Loss}
Different from the monocular depth estimation task, MHE datasets may contain images captured at diverse camera poses, e.g., different camera heights. Due to the fact that the range of object heights may vary greatly for different camera poses, as shown in Fig. \ref{ahn_show}, it will be difficult to learn consistent deep representations for the MHE model across different camera poses. In this situation, better deep representations can be learned by training MHE models with consideration to relative height relationships. Thus, during the training stage, the loss functions consist of two components. The first loss term is the regular height map regression loss, which is defined with a pixel-wise MSE loss. The second loss term is the scale-invariant training loss between different pixel pairs. Let $\bm{y}_h$ denote the ground truth height map, and $\hat{\bm{y}_h}$ be the predicted height map. Then the final loss function can be defined as
\begin{equation}
\begin{split}
    \mathcal{L}=\mathcal{L}_{mse}(\hat{\bm{y}}_{h},\bm{y}_{h}) +\mathcal{L}_{rh}(\hat{\bm{y}}_{hi},{\bm{y}}_{hj}),
\end{split}
\end{equation}
where $\mathcal{L}_{mse}$ denotes the MSE loss function. $\mathcal{L}_{rh}$ represents the relative height consistency loss. In the following part, we introduce three different implementations of the scale-invariant loss term including $\mathcal{L}_{si}$, $\mathcal{L}_{r}$, and $\mathcal{L}_{msg}$. These loss functions are initially proposed for the MHE task; in this work, we adapt them for the MHE task for performance comparison. 

To handle the varying scale problem in training depth estimation models, Eigen et al. \cite{Eigen2014Depth} designed a scale-invariant loss
\begin{equation}
\small
\begin{split}
\mathcal{L}_{si}\left(\hat{\bm{y}_h}, {\bm{y}_h}\right) &=\frac{1}{n} \sum_{i} R_{i}^{2}-\frac{1}{n^{2}}\left(\sum_{i} R_{i}\right)^{2},
\end{split}
\end{equation}
where $R_i$ is the difference between the prediction and ground truth at pixel $i$, and $R_i=\bm{y}_{hi}- \hat{\bm{y}_{hi}}$. Note that for MHE, we use the original height value instead of the log space, as a large portion of the height values are zero.

Chen et al. \cite{chen2016single} proposed to analyze the ordinal relations and enforce the model to learn relative depth:
\begin{equation}
\mathcal{L}_{r} = \begin{cases}\log \left(1+\exp \left(-\hat{\bm{y}_{i_{k}}}+\hat{\bm{y}_{j_{k}}}\right)\right), & r_{k}=+1 \\ \log \left(1+\exp \left(\hat{\bm{y}_{i_{k}}}-\hat{\bm{y}_{j_{k}}}\right)\right), & r_{k}=-1 \\ \left(\hat{\bm{y}_{i_{k}}}-\hat{\bm{y}_{j_{k}}}\right)^{2}, & r_{k}=0\end{cases}.
\end{equation}
This relative constraint loss $\mathcal{L}_r$ encourages the predicted depth map to agree with the ground-truth ordinal relations.

Ranftl et al. \cite{ranftl2019towards} proposed to use gradient matching loss \cite{wang2019web} to train the depth estimation models in the zero-shot cross-dataset transfer setting. In this work, the multiscale gradient matching loss is adapted to the MHE task by
\begin{equation}
\label{msg}
\mathcal{L}_{msg}\left(\hat{\bm{y}_h}, {\bm{y}_h}\right)=\frac{1}{M} \sum_{k=1}^{K} \sum_{i=1}^{M}\left(\left|\nabla_{x} R_{i}^{k}\right|+\left|\nabla_{y} R_{i}^{k}\right|\right),
\end{equation}
where $R_i^k$ is the difference between the predicted height map and the ground truth height map at scale $k$. The number of pixels in a predicted height map is denoted by $M$. In this work, four different scales (${1}, \frac{1}{2}, \frac{1}{4}, \frac{1}{8}$) are used. We also combine the gradient matching loss $\mathcal{L}_{msg}$ with a standard MSE loss to form the final training loss.

\begin{table}
\caption{Number of training samples for few-shot cross-dataset transfer}
\scalebox{0.8}{
\begin{tabular}{c|cc|c|c}
\hline \hline
\multirow{2}{*}{Datasets} & \multicolumn{2}{c|}{Training(\#images)} & Validation (\#images) & Test (\#images)\\ \cline{2-5} 
                          & \multicolumn{1}{c|}{1\%}   & 5\%  & ---       & ---        \\ \hline
ARG                       & \multicolumn{1}{c|}{16}    & 80   & 200     & 725      \\ \hline
ATL                       & \multicolumn{1}{c|}{4}     & 20   & 200     & 304      \\ \hline
JAX                       & \multicolumn{1}{c|}{6}     & 30   & 200     & 498      \\ \hline
OMA                       & \multicolumn{1}{c|}{12}    & 60   & 200     & 596      \\ \hline
AHN                       & \multicolumn{1}{c|}{50}    & 250  & 200     & 3871     \\ \hline \hline
\end{tabular}
}
\label{samples}
\end{table}
\begin{table}[]
	\centering
	\caption{Comparison results of different methods on the GTAH dataset. The best and second-best results are in \textcolor{blue}{blue} and \textcolor[rgb]{0,0.7,0.2}{green}.}
	\scalebox{0.85}{
	\begin{tabular}{c|c|c|c|c}
		\hline \hline
		\multirow{2}{*}{\textbf{Method}} & \multicolumn{4}{c}{\textbf{Height Estimation Metrics}}    \\ \cline{2-5}
		& \textbf{MAE} & \multicolumn{1}{l|}{\textbf{RMSE}} & \multicolumn{1}{l|}{\textbf{SI-RMSE}} & \multicolumn{1}{l}{\textbf{MSGE}} \\ \hline
		U-Net (ResNet34)\cite{christie2020geocentricpose}         & 4.860            & 6.731                                  & 39.511                                    & 3.357   \\ \hline
		Adabins (ResNet50) \cite{christie2020geocentricpose}         & 3.651            & 5.552                                  & 28.125                                    & 2.957   \\ \hline
		DenseViT \cite{christie2020geocentricpose}         & 3.253           & 4.898                                 & 23.327                                    & 2.753  \\ \hline
		SwinUper \cite{liu2021swin}          & 2.946            & 4.628                                  & 21.169                                    & 2.358   \\ \hline\hline
		SwinUper + $\mathcal{L}_{msg}$\cite{ranftl2019towards}              & \textcolor[rgb]{0,0.7,0.2}{2.943}            & \textcolor[rgb]{0,0.7,0.2}{4.614}                                  & \textcolor[rgb]{0,0.7,0.2}{20.754}                           & \textcolor{blue}{2.277}    \\ \hline
		SwinUper + $\mathcal{L}_{si}$\cite{Eigen2014Depth}               & 2.958            & 4.624                                  & {20.855}                                    & 2.392       \\ \hline
		SwinUper + $\mathcal{L}_{r}$\cite{chen2016single}           & 2.995            & 4.683                                  & 21.302                                    & 2.371    \\ \hline \hline
		SwinUper + SDC (Ours)                       & \textcolor{blue}{2.928}   & \textcolor{blue}{4.562}                         & \textcolor{blue}{19.990}                                    & \textcolor[rgb]{0,0.7,0.2}{2.336}         \\ \hline \hline
	\end{tabular}}
\label{table::GTAH_comparison}
\end{table}
\section{Experiments}
\label{experiments}
This section begins by introducing the experimental settings. Then, the evaluation metrics are defined in brief. Finally, the few-shot synthetic-to-real transfer experiments are conducted, based on the existing deep semantic models and the proposed method.

\subsection{Experimental Settings}
In this section, a series of experiments are set up to evaluate the transferability of MHE models comprehensively. 

(1) \textbf{\emph{Benchmark experiments on the GTAH dataset.}} Extensive experiments are conducted on GTAH to compare the effectiveness of different existing deep architectures,  relative height loss functions, and the proposed method in this work.

(2) \textbf{\emph{Few-shot cross-dataset transfer experiments.}} Experiments under the few-shot cross-dataset setting are performed to examine the transfer performance from the GTAH to real datasets when only a few annotated target samples are available. In addition, to better understand the effect of the proposed SDC module, we visualize and analyze the module's scale-adaptive ability for obtaining the adaptive spatial context.

(3) \textbf{\emph{Pre-training comparison experiments.}} Finally, to verify the superiority of the GTAH to ImageNet for pre-training the MHE models, their training losses and visualization of model weight distribution are provided intuitively.

\subsection{Implementation Details}
All the deep models are implemented in PyTorch. For the GTAH dataset, 100 epochs are used to train the deep models used for transfer learning in this work. For the CNN-based U-Net model with the ResNet-34 backbone, we use the code\footnote{\url{https://github.com/pubgeo/monocular-geocentric-pose}} from \cite{christie2020geocentricpose}. Adam \cite{kingma2014adam} is used for optimizing the model with an initial learning rate of 1e-4. For the SwinUper backbone, the tiny version of the Swin Transformer (Swin-T) is used and the UperNet \cite{xiao2018unified} is used for the decoder. The optimizer AdamW \cite{loshchilov2018fixing} is used with an initial learning rate of 6e-5 for training all the Transformer-based deep models. Adabins \cite{bhat2021adabins} and DenseViT \cite{ranftl2021vision} are state-of-the-art MDE models selected for performance comparison on the proposed GTAH dataset. The detailed hyperparameters for model training can be found in the publicly available code.\footnote{\url{https://github.com/EarthNets/3D-Understanding}}

In the few-shot cross-dataset transfer experiments, we randomly select 1\% and 5\% of the training data for each of the five real-world datasets (cities), as presented in Table \ref{samples}. Then 15 epochs are used to fine tune the deep models initialized with ImageNet or GTAH pre-trained parameters. Finally, for each dataset, the full test set is used for evaluating the fine-tuned models. Other details of the proposed method can be found in the publicly available code.

\subsection{Evaluation Metrics}
Different from the MDE task, the area with height value 0 accounts for a large percentage of the image. To evaluate the effectiveness of the proposed methods, we propose to use four metrics on each dataset for performance evaluation: Mean Absolute Error (MAE), Root Mean Squared Error (RMSE), Scale-Invariant RMSE (SI-RMSE) \cite{Eigen2014Depth} and Multi-scale Gradient Error (MSG). Of these, MAE and RMSE are measurements that are widely used for the evaluation of regression tasks. MAE, defined as $\text { MAE }=1 / n * \sum\left|y_{i}-\hat{y}_{i}\right|$, is used to measure the mean absolute difference between the predicted height values and the ground truth values in the whole dataset. RMSE, defined as $\text { RMSE }=\sqrt{\Sigma\left(y_{i}-\hat{y}_{i}\right)^{2} / n}$, is more sensitive to large height values. We also propose to utilize SI-RMSE and MSGE to measure the relative relationships in the predicted height maps. SI-RMSE is defined as
\begin{equation}
\text{SI-RMSE} =\frac{1}{n} \sum_{i} R_{i}^{2}-\frac{1}{n^{2}}\left(\sum_{i} R_{i}\right)^{2}.
\end{equation}
 For computing the multi-scale gradient matching error, we adopt the same formula as in Eq. \ref{msg}:
\begin{equation}
\text{MSGE} =\frac{1}{M} \sum_{k=1}^{K} \sum_{i=1}^{M}\left(\left|\nabla_{x} R_{i}^{k}\right|+\left|\nabla_{y} R_{i}^{k}\right|\right).
\end{equation}

Compared with MAE and RMSE, the metrics SI-RMSE and MSGE are more concerned with the correctness of relative relationships in height maps, which are useful complementary metrics for the evaluation of transferable MHE models.

\subsection{Experiments on GTAH Datasets}
\label{sec:GTAH_comparison}
To verify the applicability and effectiveness of the proposed GTAH dataset and evaluate the proposed SDC module fairly, eight experiments are conducted on GTAH for performance comparison.
First, following the work in \cite{christie2020geocentricpose}, the U-Net \cite{ronneberger2015u} architecture with a CNN-based feature extraction backbone (ResNet-34) is adopted as a CNN-based baseline model for MHE. Then, Adabins \cite{bhat2021adabins} with ResNet-50 backbone and DenseViT \cite{ranftl2021vision} with Vision Transformer backbone are selected as state-of-the-art MDE methods for performance comparison. Furthermore, Swin Transformer is used as a Transformer-based feature extraction backbone for height estimation, which can be viewed as another baseline. To further explore the influence of the relative height loss functions on MHE, three different types of loss functions, $\mathcal{L}_{msg}$, $\mathcal{L}_{si}$, and $\mathcal{L}_{r}$, are added to constrain the relative relationship between pair-wise pixels. Experimental results are provided in Table \ref{table::GTAH_comparison}.
\begin{table}
	\centering
	\caption{Quantitative results of SwinUper+SDC (GTAH) for Zero-shot cross-dataset transfer from GTAH to Real Cities.}
	\scalebox{0.98}{
		\begin{tabular}{c|c|c|c|c}
			\hline \hline
			{Datasets} & {MAE} & {SI-RMSE} & RMSE & MSGE\\ \cline{2-5} 
			ARG                       & 5.42	&39.61	&9.38	&7.54      \\ \hline
			ATL                       & 17.6	&156.7	&14.57	&13.05      \\ \hline
			JAX                       & 9.43	&71.58	&11.92	&7.13      \\ \hline
			OMA                       & 4.73	&51.76	&8.35	&5.41      \\ \hline
			AHN                       & 2.66	&22.18	&6.24	&5.86     \\ \hline \hline
		\end{tabular}
	}
	\label{zerotoreal}
\end{table}
When comparing the results of the two different baseline methods: U-Net and Swin Transformer, it is clear that the Transformer-based model significantly outperforms U-Net. Such results indicate that under the full-supervision setting within an unbiased dataset, Transformer architecture can be more effective on the MHE task, benefiting from its excellent context modeling capability. Based on the Swin Transformer, the performance of $\mathcal{L}_{msg}$ is superior to the other two losses $\mathcal{L}_{si}$ and $\mathcal{L}_{r}$, which reveals that a reasonable relative height constraint is useful for improving the performance of MHE. 
% Please add the following required packages to your document preamble:
% \usepackage{multirow}
\begin{table*}[]
	\small
	\centering
	\caption{Experimental results on the \textcolor{red}{AHN dataset} in the few-shot cross-dataset transfer setting. The results of using 1\% and 5\% training data are reported. The best results are in \textcolor{blue}{blue}, and the second-best ones are in \textcolor[rgb]{0,0.7,0.2}{green}.}
	\scalebox{0.98}{
		\begin{tabular}{c|c|c|c|c|c|c|c|c}
			\hline \hline
			\multirow{2}{*}{Methods} & \multicolumn{4}{c|}{\textbf{Height Estimation Metrics (1\% Training)}}                   & \multicolumn{4}{c}{\textbf{Height Estimation Metrics (5\% Training)}}    \\ \cline{2-9} 
			& \textbf{MAE}         & \textbf{SI-RMSE}        & \textbf{RMSE}     & \textbf{MSGE}   & \textbf{MAE} & \textbf{SI-RMSE} & \textbf{RMSE}     & \textbf{MSGE}   \\ \hline
			U-Net (ImageNet)\cite{christie2020geocentricpose}      & 2.348      & 8.739     & 5.759    & 1.620   & 2.313  & 9.988   & 5.600      & 1.564     \\ \hline
			U-Net (GTAH)\cite{christie2020geocentricpose}              & 2.289          & 8.334       & 5.673          & 1.536    & 2.256  & 9.716   & 5.632    & 1.520 \\ \hline
			Adabins (ImageNet)\cite{bhat2021adabins}          & 2.316    & 8.707      & 5.677          & 1.583     & 2.275  & 9.531   & 5.541      & 1.537      \\ \hline
			Adabins (GTAH)\cite{bhat2021adabins}              & 2.273      & 8.345     & 5.568         & 1.530   & 2.136  & 9.232   &  5.461         & 1.499          \\ \hline
			DenseViT (ImageNet)\cite{ranftl2021vision}          & 2.524          & 8.873         & 5.907      & 1.627    & 2.415  & 9.990  & 5.769      & 1.617  \\ \hline
			DenseViT (GTAH)\cite{ranftl2021vision}              & 2.439         & 8.633       & 5.986       & 1.616   & 2.314 & 9.217   & 5.640      & 1.536     \\ \hline
			SwinUper (ImageNet)\cite{liu2021swin}           & 2.394   & 9.002      & 5.694       & 1.724      & 2.104  & 9.297   & 5.585         & 1.435        \\ \hline
			SwinUper (GTAH)\cite{liu2021swin}               & \textcolor[rgb]{0,0.7,0.2}{2.156}   &\textcolor[rgb]{0,0.7,0.2}{8.303} & \textcolor{blue}{5.564}          & \textcolor[rgb]{0,0.7,0.2}{1.455} & \textcolor[rgb]{0,0.7,0.2}{2.107}  & \textcolor[rgb]{0,0.7,0.2}{9.021}   & 5.569          & \textcolor[rgb]{0,0.7,0.2}{1.426}          \\ \hline
			SwinUper+$\mathcal{L}_{msg}$ (GTAH)\cite{ranftl2019towards}    & 2.160     & 8.343       & {5.584}          & 1.467      & 2.112  & 9.040   & \textcolor[rgb]{0,0.7,0.2}{5.569}          & 1.430          \\ \hline
			SwinUper+SDC (GTAH) (Ours)      & \textcolor{blue}{2.117}          & \textcolor{blue}{8.285}          & \textcolor[rgb]{0,0.7,0.2}{5.530} & \textcolor{blue}{1.441}          & \textcolor{blue}{2.001}  & \textcolor{blue}{8.981}   & \textcolor{blue}{5.439} & \textcolor{blue}{1.417} \\ \hline \hline
		\end{tabular}
	}
\label{table:few_AHN}
\end{table*}

When integrating the proposed SDC module into the Swin Transformer, a notable performance gain is obtained, which verifies the effectiveness of the proposed SDC module for MHE due to its adaptive context modeling ability. It is worth noting that there is great potential to combine $\mathcal{L}_{msg}$ with our proposed SDC, which may further boost MHE performance.

\begin{figure}
	\begin{center}
		\includegraphics[width=0.425\textwidth]{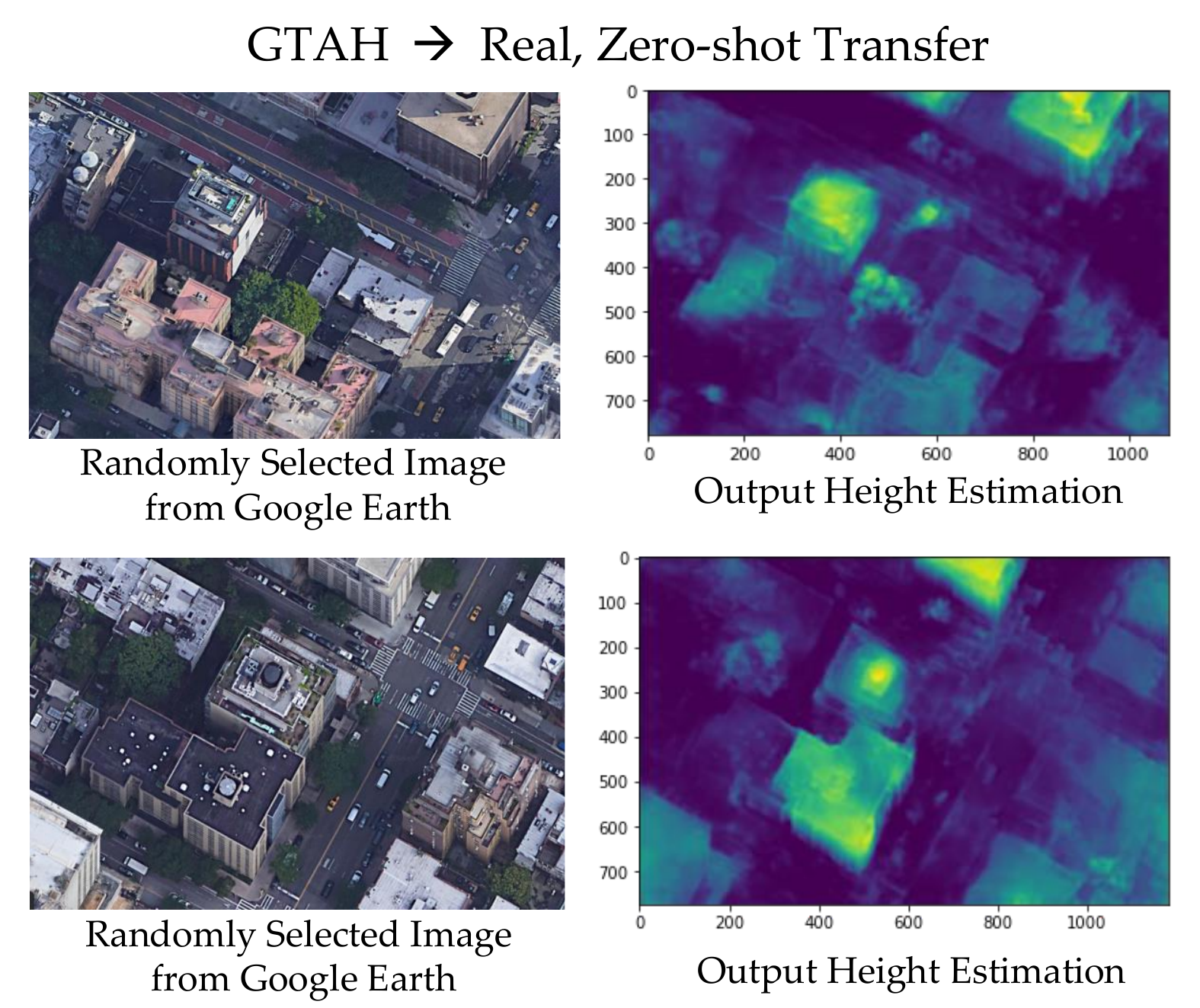}
	\end{center}
	\caption{Visualization of some height estimation samples on a few RGB images randomly selected from Google Earth. The MHE model is pre-trained on the GTAH dataset, then directly used to predict the normalized height maps of these images without fine-tuning.}
	\label{fig:ZeroToReal}
\end{figure}

To further explore the generalization of the proposed method, we directly use the proposed method ``SwinUper+SDC" that is pre-trained on GTAH to predict the height of real-world images. Note that these images are randomly selected from Google Earth Map, as illustrated in Fig. \ref{fig:ZeroToReal}. We can see that the pre-trained model can reasonably estimate the height maps of these images, especially for the buildings with clear geometric information.

\subsection{Experiments on Few-shot Cross-dataset Transfer}
Few-shot learning has been studied heavily for image classification and semantic segmentation. However, for the dense regression task like MHE, there is still a lack of research on few-shot cross-dataset transfer. In this work, we fill in this gap by conducting cross-dataset transfer experiments from GTAH to the five real-world datasets under the few-shot setting. Before the few-shot cross-dataset transfer setting, we first conduct zero-shot transfer experiments to show the MHE results on real-city datasets using the GTAH pre-trained weights with no finetuning process. The results in Table \ref{zerotoreal} reveal that the MHE performance in the zero-shot transfer setting is poor due to significant domain shifts.

In the few-shot setting, for each dataset, only 1\% or 5\% of the training samples are randomly sampled for fast fine-tuning. As presented in Table \ref{samples}, for 1\% setting, only less than 100 images are used for the model finetuning. The experimental results of AHN, JAX, OMA, ATL, and ARG are shown in Table \ref{table:few_AHN}, Table \ref{table:few_JAX}, Table \ref{table:few_OMA}, Table \ref{table:few_ATL}, and Table \ref{table:few_ARG}, respectively.

\vspace{2em}

\begin{table*}[]
	\small
	\centering
	\caption{Experimental results on the \textcolor{red}{JAX dataset} in the few-shot cross-dataset transfer setting. The results of using 1\% and 5\% training data are reported. The best results are in \textcolor{blue}{blue}, and the second-best ones are in \textcolor[rgb]{0,0.7,0.2}{green}.}
	\scalebox{0.98}{
	\begin{tabular}{c|c|c|c|c|c|c|c|c}
		\hline \hline
		\multirow{2}{*}{Methods} & \multicolumn{4}{c|}{\textbf{Height Estimation Metrics (1\% Training)}}                    & \multicolumn{4}{c}{\textbf{Height Estimation Metrics (5\% Training)}}                    \\ \cline{2-9} 
		& \textbf{MAE}         & \textbf{SI-RMSE}         & \textbf{RMSE}     & \textbf{MSGE}   & \textbf{MAE}         & \textbf{SI-RMSE}         & \textbf{RMSE}     & \textbf{MSGE}   \\ \hline
		U-Net (ImageNet)\cite{christie2020geocentricpose}          & 7.403        & 54.166       & 9.891         & 4.038         & 5.738          & 40.100          & 6.807         & 3.204          \\ \hline
		U-Net (GTAH)\cite{christie2020geocentricpose}              & 6.297         & 49.096          & 9.345       & 3.703          & 5.691         & 35.678         & 6.703         & 3.181          \\ \hline
		Adabins (ImageNet)\cite{bhat2021adabins}          & 6.510         & 48.972         & 9.001        & 3.970         & 5.22 & 35.239   & 6.592        & 3.202          \\ \hline
		Adabins (GTAH)\cite{bhat2021adabins}              & 5.779          & 40.335       & 7.327          & 3.522         & 5.082  & 27.418   & \textcolor[rgb]{0,0.7,0.2}{6.390}          & 3.092          \\ \hline
		DenseViT (ImageNet)\cite{ranftl2021vision}          & 7.165          & 50.660         & 9.777          & 3.975         & 5.676  & 35.543   & 6.534          & 3.144          \\ \hline
		DenseViT (GTAH)\cite{ranftl2021vision}              & 5.936         & 40.695          & 8.667          & 3.708          & 5.648  & 29.517   & 6.768          & 3.103          \\ \hline
		SwinUper (ImageNet)\cite{liu2021swin}           & 6.281          & 43.410          & 6.984           & 3.328          & 5.694          & 31.480          &  6.950         & 3.060          \\ \hline
		SwinUper (GTAH)\cite{liu2021swin}           & \textcolor[rgb]{0,0.7,0.2}{5.597}          & \textcolor[rgb]{0,0.7,0.2}{40.314}          & \textcolor[rgb]{0,0.7,0.2}{6.600}         & \textcolor[rgb]{0,0.7,0.2}{3.324}          & \textcolor[rgb]{0,0.7,0.2}{4.739}          & \textcolor[rgb]{0,0.7,0.2}{25.276}          & 6.602          & \textcolor[rgb]{0,0.7,0.2}{2.920}          \\ \hline
		SwinUper+$\mathcal{L}_{msg}$ (GTAH)\cite{ranftl2019towards}     & 5.645          & 40.719          &{6.818}          & 3.420          & 4.828          & 26.054         & {6.632}          & 2.998          \\ \hline
		SwinUper+SDC (GTAH) (Ours)       & \textcolor{blue}{5.425} & \textcolor{blue}{37.773} & \textcolor{blue}{6.530} & \textcolor{blue}{3.136} & \textcolor{blue}{4.576} & \textcolor{blue}{23.506} & \textcolor{blue}{6.211} & \textcolor{blue}{2.799} \\ \hline \hline
	\end{tabular}
}
\label{table:few_JAX}
\vspace{2em}

%\begin{table*}[]
	\small
	\centering
	\caption{Experimental results on the \textcolor{red}{OMA dataset} in the few-shot cross-dataset transfer setting. The results of using 1\% and 5\% training data are reported. The best results are in \textcolor{blue}{blue}, and the second-best ones are in \textcolor[rgb]{0,0.7,0.2}{green}.}
	\scalebox{0.98}{
		\begin{tabular}{c|c|c|c|c|c|c|c|c}
			\hline \hline
			\multirow{2}{*}{Methods} & \multicolumn{4}{c|}{\textbf{Height Estimation Metrics (1\% Training)}}                    & \multicolumn{4}{c}{\textbf{Height Estimation Metrics (5\% Training)}}                   \\ \cline{2-9} 
			& \textbf{MAE}         & \textbf{SI-RMSE}         & \textbf{RMSE}     & \textbf{MSGE}   & \textbf{MAE}         & \textbf{SI-RMSE}        & \textbf{RMSE}     & \textbf{MSGE}   \\ \hline
			U-Net (ImageNet)\cite{christie2020geocentricpose}          & 3.755          & 17.682          & 6.399          & 2.031          & 3.475          & 19.402         & 6.156          & 1.925          \\ \hline
			U-Net (GTAH)\cite{christie2020geocentricpose}              & 3.572          & 16.452          & 6.388          & 2.009          & 3.439          & 14.815         & 6.028          & 1.873          \\ \hline
			Adabins (ImageNet)\cite{bhat2021adabins}          & 3.431          & 17.630        & 6.125          &  1.915      & 3.248  & 18.977   & 6.101          & 1.876          \\ \hline
			Adabins (GTAH)\cite{bhat2021adabins}              & 3.334          & 14.725         & 5.291          & \textcolor[rgb]{0,0.7,0.2}{1.887}        & 2.913  & 13.815   & 4.838          & 1.618         \\ \hline
			DenseViT (ImageNet)\cite{ranftl2021vision}          & 3.665          & 17.679          & 6.203          & 1.974          & 2.593  & 18.379   & 5.913          & 1.872        \\ \hline
			DenseViT (GTAH)\cite{ranftl2021vision}              & 3.546          & 15.257          & 5.619          & 1.898          & 3.250  & 11.009   & 5.490          & 1.650         \\ \hline
			SwinUper (ImageNet)\cite{liu2021swin}           & 3.780          & 16.350          & \textcolor[rgb]{0,0.7,0.2}{4.549} & 1.980 & 3.577          & 14.757         & 4.465          & 1.786          \\ \hline
			SwinUper (GTAH)\cite{liu2021swin}               & \textcolor[rgb]{0,0.7,0.2}{3.318}          & \textcolor[rgb]{0,0.7,0.2}{14.247}          & 4.600          & {1.969}          & \textcolor[rgb]{0,0.7,0.2}{2.800}          & \textcolor[rgb]{0,0.7,0.2}{9.956}          & \textcolor[rgb]{0,0.7,0.2}{4.439}          & \textcolor[rgb]{0,0.7,0.2}{1.582}          \\ \hline
			SwinUper+$\mathcal{L}_{msg}$ (GTAH)\cite{ranftl2019towards}     & 3.411          & 14.826         & {4.572}          & 2.109          & 2.896          & 10.625         & 4.401          & 1.604         \\ \hline
			SwinUper+SDC (GTAH) (Ours)       & \textcolor{blue}{3.237} & \textcolor{blue}{13.714} & \textcolor{blue}{4.480} & \textcolor{blue}{1.866} & \textcolor{blue}{2.735} & \textcolor{blue}{9.519} & \textcolor{blue}{4.238} & \textcolor{blue}{1.536} \\ \hline \hline
		\end{tabular}
	}
\label{table:few_OMA}
\vspace{2em}

%\begin{table*}[]
	\small
	\centering
	\caption{Experimental results on the \textcolor{red}{ATL dataset} in the few-shot cross-dataset transfer setting. The results of using 1\% and 5\% training data are reported. The best results are in \textcolor{blue}{blue}, and the second-best ones are in \textcolor[rgb]{0,0.7,0.2}{green}.}
	\scalebox{0.98}{
		\begin{tabular}{c|c|c|c|c|c|c|c|c}
			\hline \hline
			\multirow{2}{*}{Methods} & \multicolumn{4}{c|}{\textbf{Height Estimation Metrics (1\% Training)}}                     & \multicolumn{4}{c}{\textbf{Height Estimation Metrics (5\% Training)}}                     \\ \cline{2-9} 
			& \textbf{MAE}         & \textbf{SI-RMSE}         & \textbf{RMSE}      & \textbf{MSGE}   & \textbf{MAE}         & \textbf{SI-RMSE}         & \textbf{RMSE}      & \textbf{MSGE}   \\ \hline
			U-Net (ImageNet)\cite{christie2020geocentricpose}          & 13.693        & 143.72     & 12.985         & 8.446         & 12.503        & 129.958      & 17.871         & 7.384          \\ \hline
			U-Net (GTAH)\cite{christie2020geocentricpose}              & 13.354         & 146.156         & 12.884          & 7.893          & 10.160         & 123.002         & 18.863          & 6.228         \\ \hline
			Adabins (ImageNet)\cite{bhat2021adabins}          & 11.711          & 122.33          & 12.082          & 8.147          & 10.589  & 119.908   & 17.221          & 7.174          \\ \hline
			Adabins (GTAH)\cite{bhat2021adabins}              & 9.520         & 106.644         & 11.921          & 7.014         & 9.072  & 88.268   & \textcolor[rgb]{0,0.7,0.2}{12.405}          & 5.579          \\ \hline
			DenseViT (ImageNet)\cite{ranftl2021vision}          & 12.791          & 135.99          & 12.618          & 8.161         & 12.480  & 129.946   & 17.809         & 7.358          \\ \hline
			DenseViT (GTAH)\cite{ranftl2021vision}              & 10.296        & 119.817         & 11.612          & 7.167          & 9.385  & 129.759   & 13.784          & 7.105         \\ \hline
			SwinUper (ImageNet)\cite{liu2021swin}           & 13.827         & 143.518       & 12.732          & 8.199          & 10.370         & 99.236         & 12.566         & 6.057          \\ \hline
			SwinUper (GTAH)\cite{liu2021swin}               & \textcolor[rgb]{0,0.7,0.2}{9.431} & \textcolor{blue}{98.307} & \textcolor[rgb]{0,0.7,0.2}{11.718}          & \textcolor[rgb]{0,0.7,0.2}{6.351} & \textcolor[rgb]{0,0.7,0.2}{7.964} & \textcolor[rgb]{0,0.7,0.2}{76.787} & {12.987}        & 4.850          \\ \hline
			SwinUper+$\mathcal{L}_{msg}$ (GTAH)\cite{ranftl2019towards}      & 9.519          & 99.490          & 11.757          & 6.392          & {7.980}          & \textcolor{blue}{77.112}          & {12.938}          & \textcolor[rgb]{0,0.7,0.2}{4.844}          \\ \hline
			SwinUper+SDC (GTAH) (Ours)        & \textcolor{blue}{9.386}          & \textcolor[rgb]{0,0.7,0.2}{98.738}        & \textcolor{blue}{11.605} & \textcolor{blue}{6.307}          & \textcolor{blue}{7.959}          & 77.427          & \textcolor{blue}{11.913} & \textcolor{blue}{4.836} \\ \hline \hline
		\end{tabular}
	}
\label{table:few_ATL}
%\end{table*}

\vspace{2em}
vim
%\begin{table*}[]
	\small
	\centering
	\caption{Experimental results on the \textcolor{red}{ARG dataset} in the few-shot cross-dataset transfer setting. The results of using 1\% and 5\% training data are reported. The best results are in \textcolor{blue}{blue}, and the second-best ones are in \textcolor[rgb]{0,0.7,0.2}{green}.}
	\scalebox{0.98}{
		\begin{tabular}{c|c|c|c|c|c|c|c|c}
			\hline \hline
			\multirow{2}{*}{Methods} & \multicolumn{4}{c|}{\textbf{Height Estimation Metrics (1\% Training)}}                    & \multicolumn{4}{c}{\textbf{Height Estimation Metrics (5\% Training)}}                   \\ \cline{2-9} 
			& \textbf{MAE}         & \textbf{SI-RMSE}         & \textbf{RMSE}     & \textbf{MSGE}   & \textbf{MAE}         & \textbf{SI-RMSE}        & \textbf{RMSE}     & \textbf{MSGE}   \\ \hline
			U-Net (ImageNet)\cite{christie2020geocentricpose}          & 3.658          & 12.561          & 8.484          & 2.497          & 3.392          & 12.168         & 8.351          & 2.398          \\ \hline
			U-Net (GTAH)\cite{christie2020geocentricpose}              & 3.611         & 13.604          & 8.546          & 2.581          & 3.091          & 10.217         & 8.121         & 2.146          \\ \hline
			Adabins (ImageNet)\cite{bhat2021adabins}          & 3.621          & 12.551        & 7.007          & 2.486         & 3.208  & 11.089   & 6.693          & 2.099          \\ \hline
			Adabins (GTAH)\cite{bhat2021adabins}              & \textcolor[rgb]{0,0.7,0.2}{3.465}          & \textcolor[rgb]{0,0.7,0.2}{11.365}          &  6.719         & 2.388          & 2.904  & 9.712   & 6.211   & 1.942          \\ \hline
			DenseViT (ImageNet)\cite{ranftl2021vision}          & 3.632          & 12.389          & 8.261         & 2.485          & 3.173  & 11.998  & 7.703          & 2.311          \\ \hline
			DenseViT (GTAH)\cite{ranftl2021vision}              & 3.520          & 11.393          &    7.452       & 2.395          & 2.972  & 10.053   & 4.997          & 1.941          \\ \hline
			SwinUper (ImageNet)\cite{liu2021swin}           & 3.742          & 12.516          & 6.340          & 2.523         & 2.894         & 9.548         & 6.107         & 2.004         \\ \hline
			SwinUper (GTAH)\cite{liu2021swin}               & {3.486}          & 11.976          & 6.260          & 2.508          & \textcolor[rgb]{0,0.7,0.2}{2.831}          & \textcolor[rgb]{0,0.7,0.2}{9.015}          & 6.048          & \textcolor[rgb]{0,0.7,0.2}{1.910}          \\ \hline
			SwinUper+$\mathcal{L}_{msg}$ (GTAH)\cite{ranftl2019towards}               & 3.497          & {11.910}          & \textcolor[rgb]{0,0.7,0.2}{6.193}          & \textcolor[rgb]{0,0.7,0.2}{2.474}          & 2.926          & 9.774          & \textcolor[rgb]{0,0.7,0.2}{6.096}          & 2.038          \\ \hline
			SwinUper+SDC (GTAH) (Ours)                & \textcolor{blue}{3.458} & \textcolor{blue}{10.467} & \textcolor{blue}{6.187} & \textcolor{blue}{2.230} & \textcolor{blue}{2.786} & \textcolor{blue}{8.883} & \textcolor{blue}{6.030} & \textcolor{blue}{1.804} \\ \hline \hline
		\end{tabular}
	}
\label{table:few_ARG}
\end{table*}

%\begin{comment}
\begin{table*}[]
	\small
	\centering
	\caption{Experimental results on the AHN dataset with different pre-training methods. The best results are in \textcolor{blue}{blue}, and the second-best ones are in \textcolor[rgb]{0,0.7,0.2}{green}.}
	\scalebox{1.0}{
		\begin{tabular}{c|c|c|c|c|c|c|c|c}
			\hline \hline
			\multirow{2}{*}{Methods} & \multicolumn{4}{c|}{\textbf{Height Estimation Metrics (1\% Training)}}                   & \multicolumn{4}{c}{\textbf{Height Estimation Metrics (5\% Training)}}    \\ \cline{2-9} 
			& \textbf{MAE}         & \textbf{SI-RMSE}        & \textbf{RMSE}     & \textbf{MSGE}   & \textbf{MAE} & \textbf{SI-RMSE} & \textbf{RMSE}     & \textbf{MSGE}   \\ \hline
			SwinUper (ImageNet)\cite{liu2021swin}           & 2.394   & 9.002      & 5.694       & 1.724      & 2.104  & 9.297   & 5.585         & 1.435        \\ \hline
            SwinUper (SSLTransformerRS)\cite{scheibenreif2022self}           & 2.347   & 8.626      & 5.715       & 1.613      & 2.161  & 9.055   & 5.674         & \textcolor[rgb]{0,0.7,0.2}{1.423}        \\ \hline
			SwinUper (GTAH)\cite{liu2021swin}               & \textcolor[rgb]{0,0.7,0.2}{2.156}   &\textcolor[rgb]{0,0.7,0.2}{8.303} & \textcolor{blue}{5.564}          & \textcolor[rgb]{0,0.7,0.2}{1.455} & \textcolor[rgb]{0,0.7,0.2}{2.107}  & \textcolor[rgb]{0,0.7,0.2}{9.021}   & \textcolor{blue}{5.569}          & {1.426}          \\ \hline
			SwinUper+SDC (GTAH) (Ours)      & \textcolor{blue}{2.117}          & \textcolor{blue}{8.285}          & \textcolor[rgb]{0,0.7,0.2}{5.530} & \textcolor{blue}{1.441}          & \textcolor{blue}{2.001}  & \textcolor{blue}{8.981}   & {5.439} & \textcolor{blue}{1.417} \\ \hline \hline
		\end{tabular}
	}
\label{table:pretrain_AHN}
\end{table*}
%\end{comment}

When we focus on the initialization strategy, it can be seen that the results of all the different models including U-Net, Adabins, DenseViT and the SwinUper pre-trained on GTAH have a dramatic superiority to those pre-trained on ImageNet. Especially for the ATL dataset with higher height distribution, ImageNet pre-trained U-Net and Swin Transformer experience a performance collapse, whereas these models pre-trained on GTAH maintain a stable performance. The results demonstrate that our proposed GTAH dataset is more suitable for MHE initialization. 

Next, the results of the Swin Transformer model pre-trained on GTAH with relative constraint loss $\mathcal{L}_{msg}$ show that the introduction of $\mathcal{L}_{msg}$ is not beneficial for improving performance in general. The reason may be that introducing the relative constraint by loss functions is not useful for improving the generalizability of the MHE model across different datasets.
In contrast, the proposed SDC module is still effective in the few-shot setting. The model ``SwinUper+SDC (GTAH)" obtains the best results on all the datasets. Compared with the baseline method ``SwinUper (GTAH)," the proposed model ``SwinUper+SDC (GTAH)" shows virtue of its overall superiority with a considerable performance gain, which verifies that the adaptive scale modeling ability is helpful for the Swin Transformer by learning an adaptive receptive-field.

To intuitively illustrate the effect of the SDC module, some visualization examples of the dynamic spatial context are provided in Fig. \ref{fig:visual_ASC}. In this figure, we can see that the low-frequency region needs a larger receptive field to acquire enough context information, while the high-frequency region only requires a relatively smaller receptive field for the MHE task. 

Some visualization examples on the five real-world datasets under the few-shot transfer setting are presented in Fig. \ref{FewV}.

\begin{figure}
	%\vspace{-0.8cm}
	\begin{center}
		\includegraphics[width=0.485\textwidth]{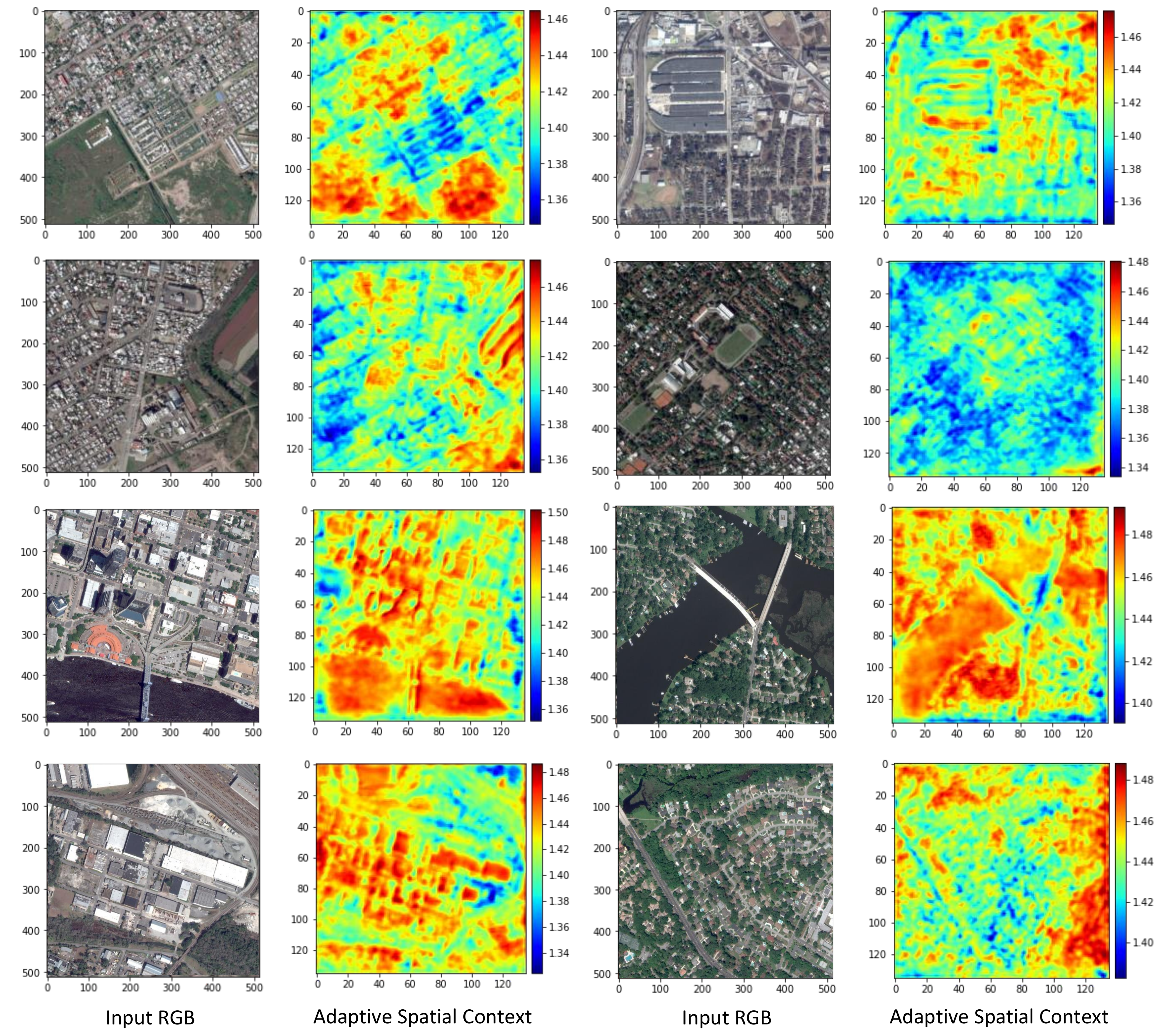}
	\end{center}
	\setlength{\belowcaptionskip}{-0.5cm}
	\setlength{\abovecaptionskip}{-0.0cm}
	
	\caption{Some visualization examples of the proposed SDC module. The SDC module can be observed to have learned to adjust the context information of different pixels in the cross-dataset transfer settings.}
	\label{fig:visual_ASC}
\end{figure}

\subsection{ImageNet Pre-training vs. GTAH Pre-training}
To further study the effectiveness of model pre-training in the cross-dataset transfer setting, we present and analyze the loss trends of the model ``SwinUper+SDC" during the few-shot model finetuning stage.
As shown in Fig. \ref{fig:train_loss}, from the loss trend we can see that models with GTAH pre-trained parameters can converge faster. Especially on the JAX, OMA, and ATL datasets, models initialized with the ImageNet pre-trained parameters are difficult to converge. 
\begin{figure*}
	%\vspace{-0.8cm}
	\begin{center}
		\includegraphics[width=0.995\textwidth]{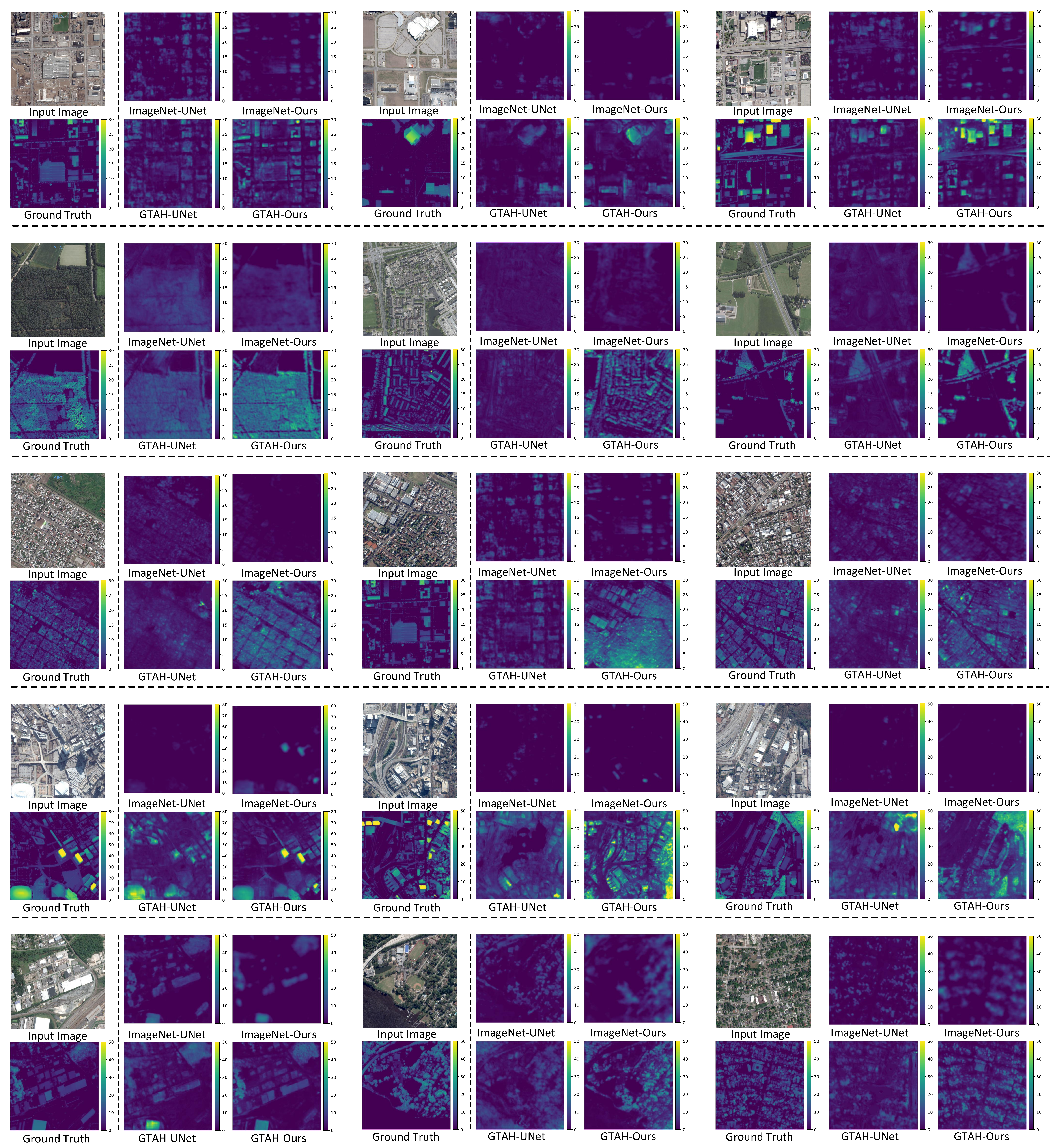}
	\end{center}
	\setlength{\belowcaptionskip}{-0.5cm}
	\setlength{\abovecaptionskip}{-0.0cm}
	\caption{Visualization results of height estimation in a few-shot \textbf{(1\% of the training data)} cross-dataset transfer setting. In this figure, the predicted height maps of five real-world datasets OMA, AHN, ARG, ATL, and JAX are shown. It can be seen that our method ``SwinUper+SDC (GTAH)'' can obtain much better results than other methods in the few-shot across-dataset transfer setting.}
	\label{FewV}
\end{figure*}
\begin{figure*}
	\centering
	\subfigure[AHN]{
		\begin{minipage}[b]{0.18\textwidth}
			\includegraphics[width=1\textwidth]{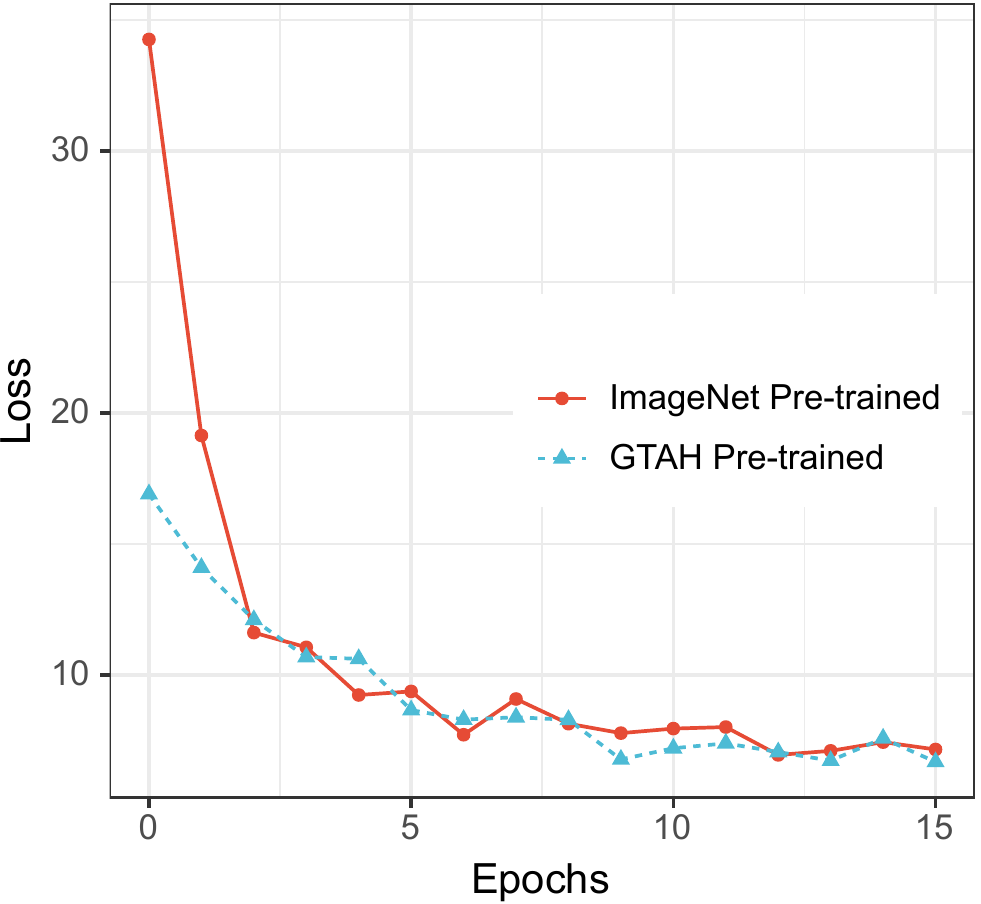} 
		\end{minipage}
		\label{fig:grid_4figs_1cap_4subcap_1}
	}
    	\subfigure[JAX]{
    		\begin{minipage}[b]{0.18\textwidth}
   		 	\includegraphics[width=1\textwidth]{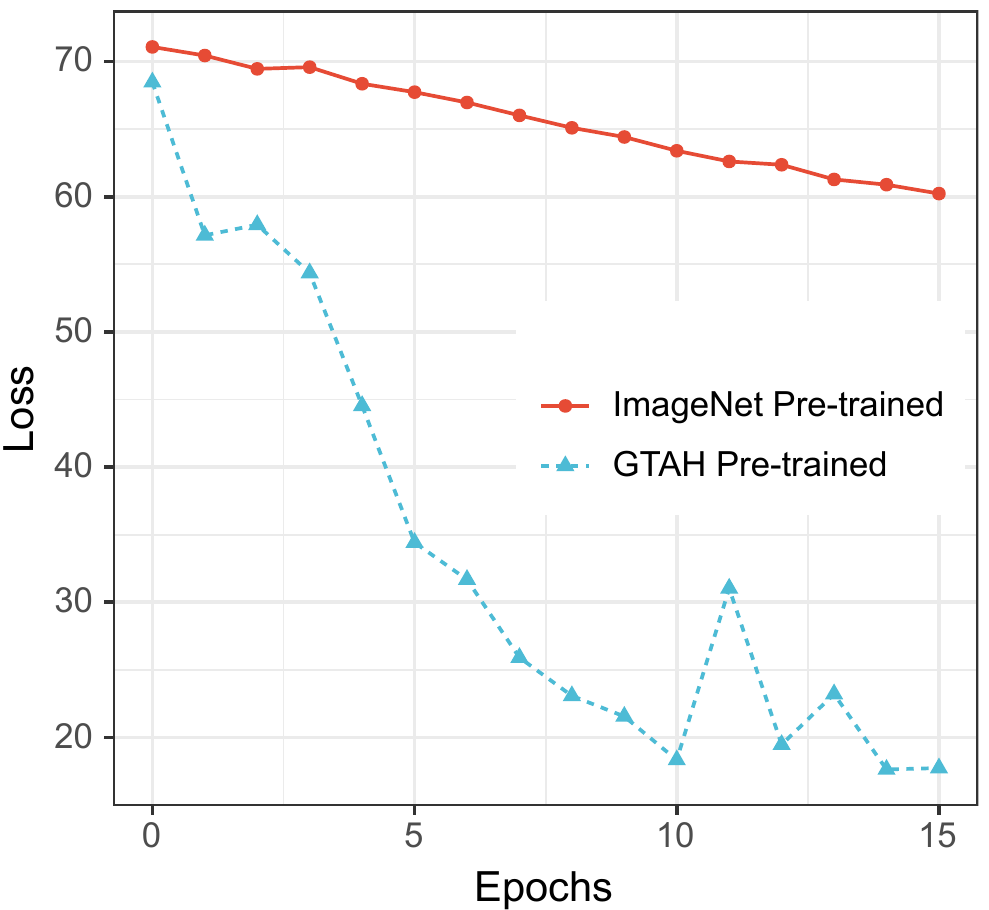}
    		\end{minipage}
		\label{fig:grid_4figs_1cap_4subcap_2}
    	}
	\subfigure[OMA]{
		\begin{minipage}[b]{0.18\textwidth}
			\includegraphics[width=1\textwidth]{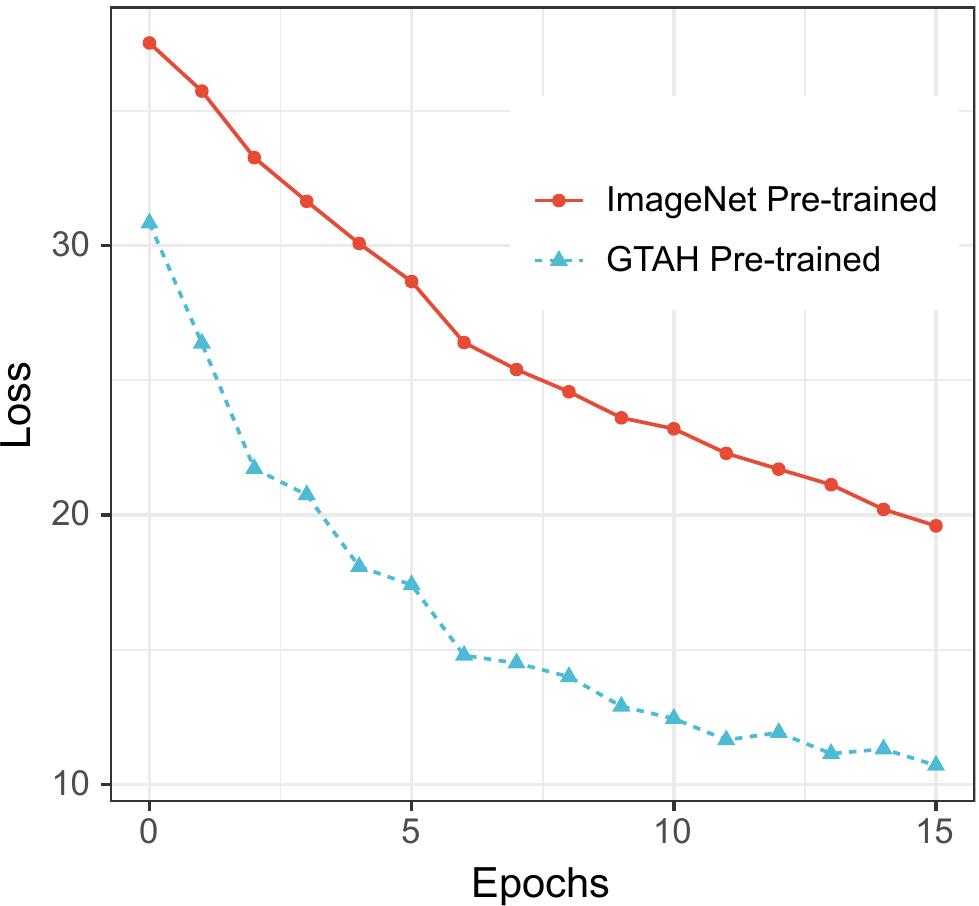} 
		\end{minipage}
		\label{fig:grid_4figs_1cap_4subcap_3}
	}
    	\subfigure[ATL]{
    		\begin{minipage}[b]{0.18\textwidth}
		 	\includegraphics[width=1\textwidth]{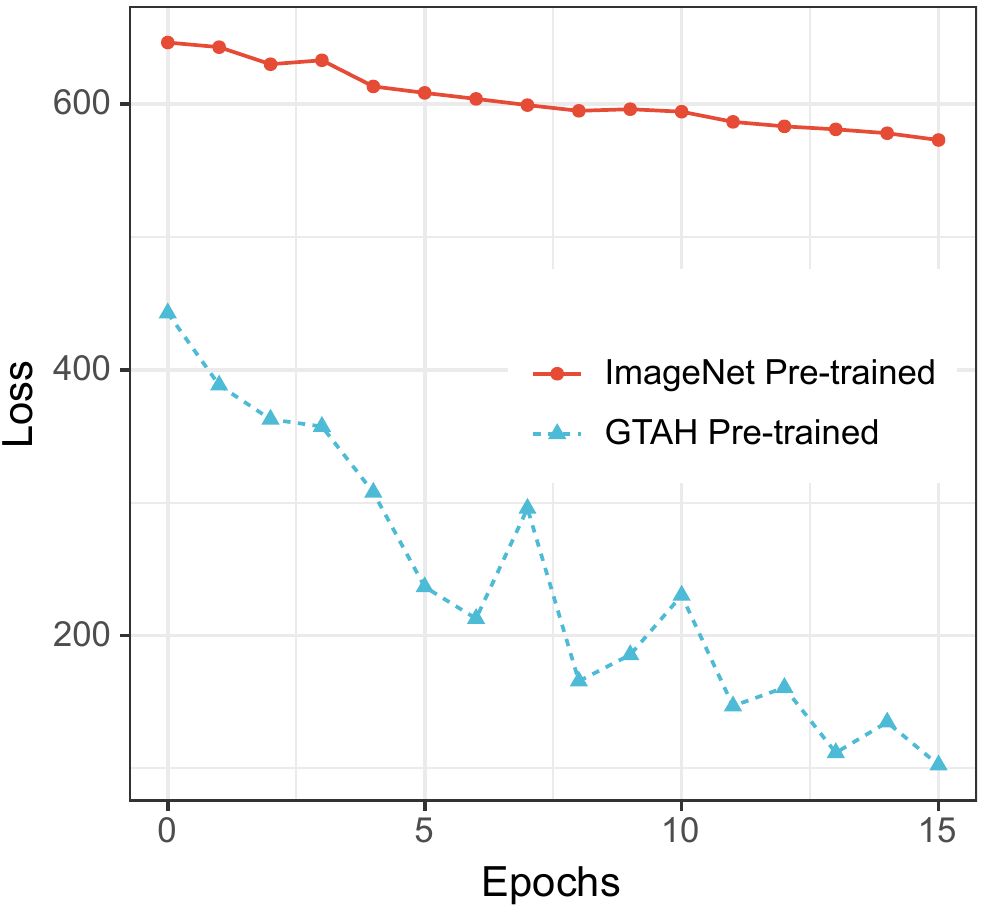}
    		\end{minipage}
		\label{fig:grid_4figs_1cap_4subcap_4}
    	}
    	\subfigure[ARG]{
    		\begin{minipage}[b]{0.18\textwidth}
		 	\includegraphics[width=1\textwidth]{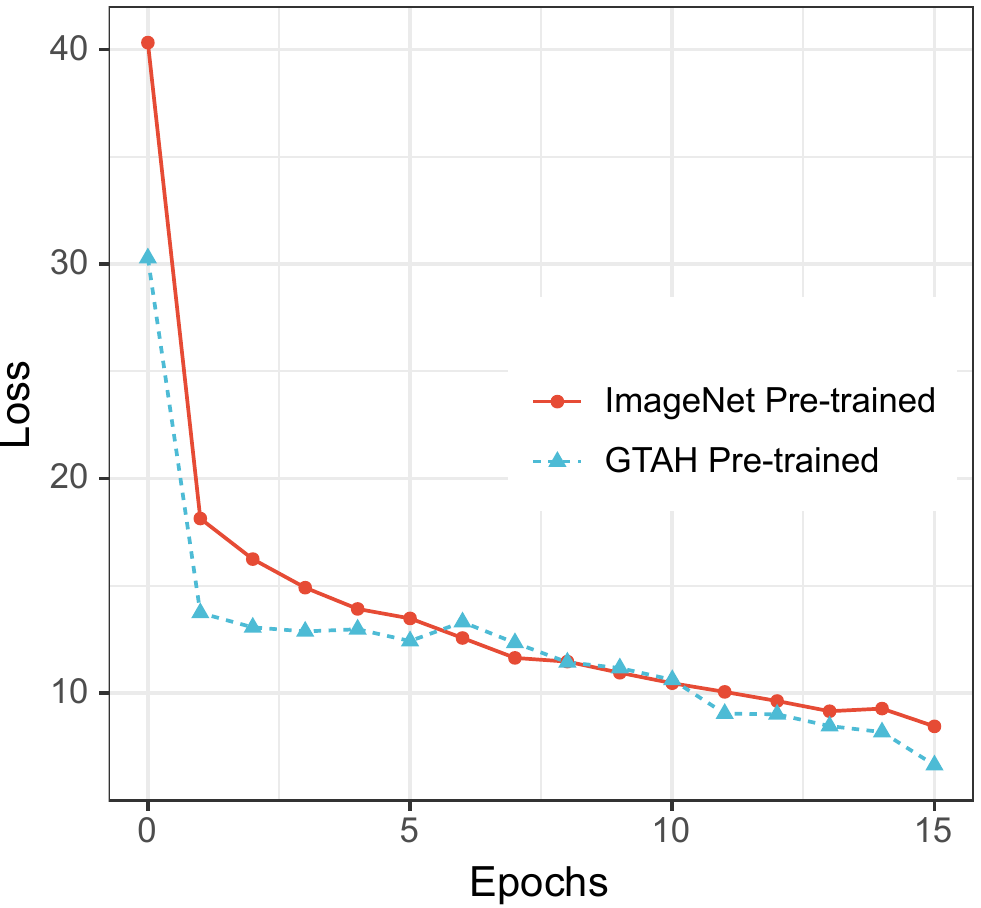}
    		\end{minipage}
		\label{fig:grid_4figs_1cap_4subcap_4}
    	}
	\caption{Training loss of the ``Swin-SDC" (Swin Transformer followed by the SDC module) on five real-world MHE datasets after being pre-trained on ImageNet and GTAH.}
	\label{fig:train_loss}
\end{figure*}

\begin{figure*}
	\centering
	\subfigure[Layer1-ImageNet]{
		\begin{minipage}[b]{0.31\textwidth}
			\includegraphics[width=1\textwidth]{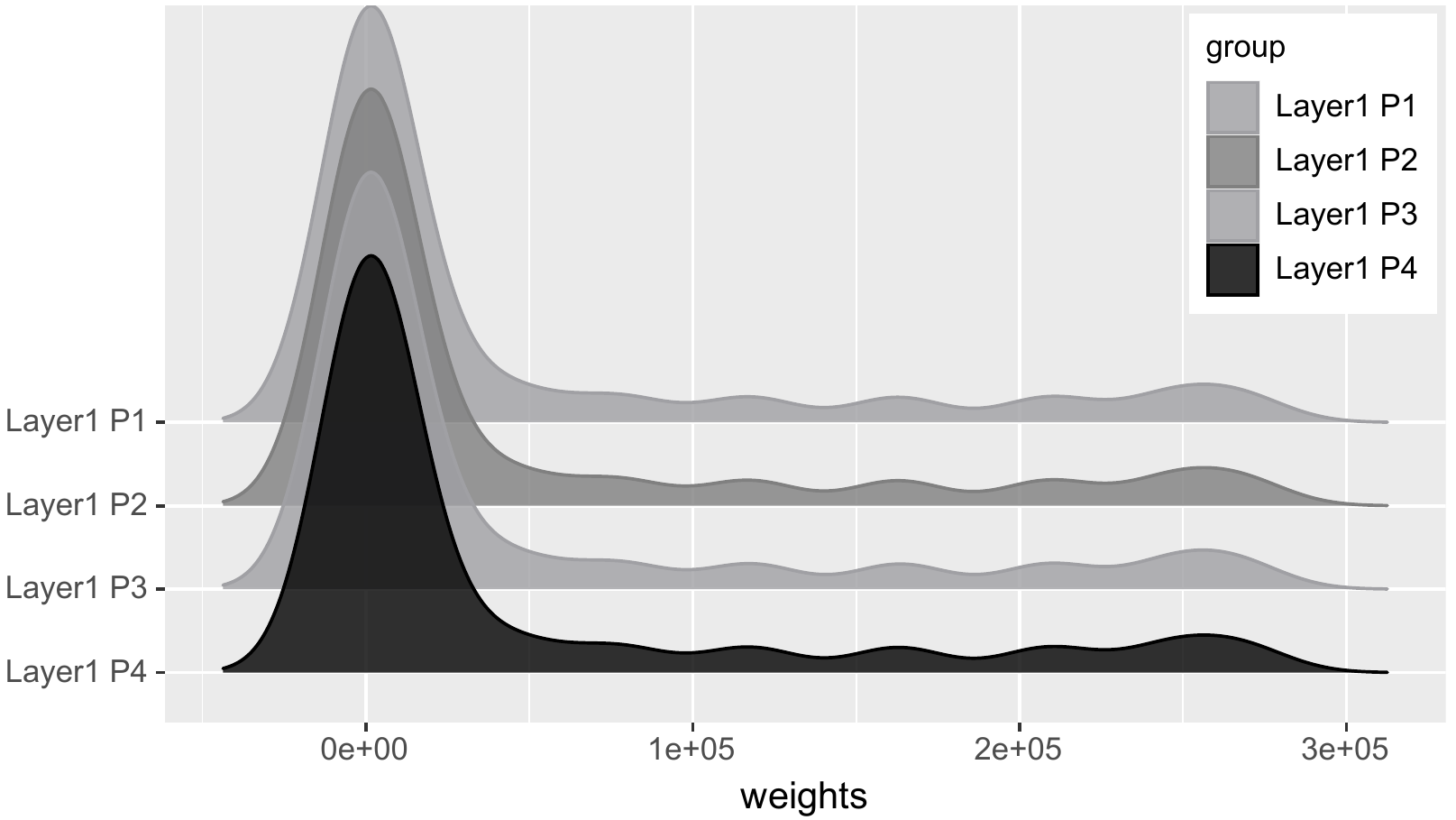} 
		\end{minipage}
		\label{fig:grid_4figs_1cap_4subcap_1}
	}
    	\subfigure[Layer1-GTAH]{
    		\begin{minipage}[b]{0.31\textwidth}
   		 	\includegraphics[width=1\textwidth]{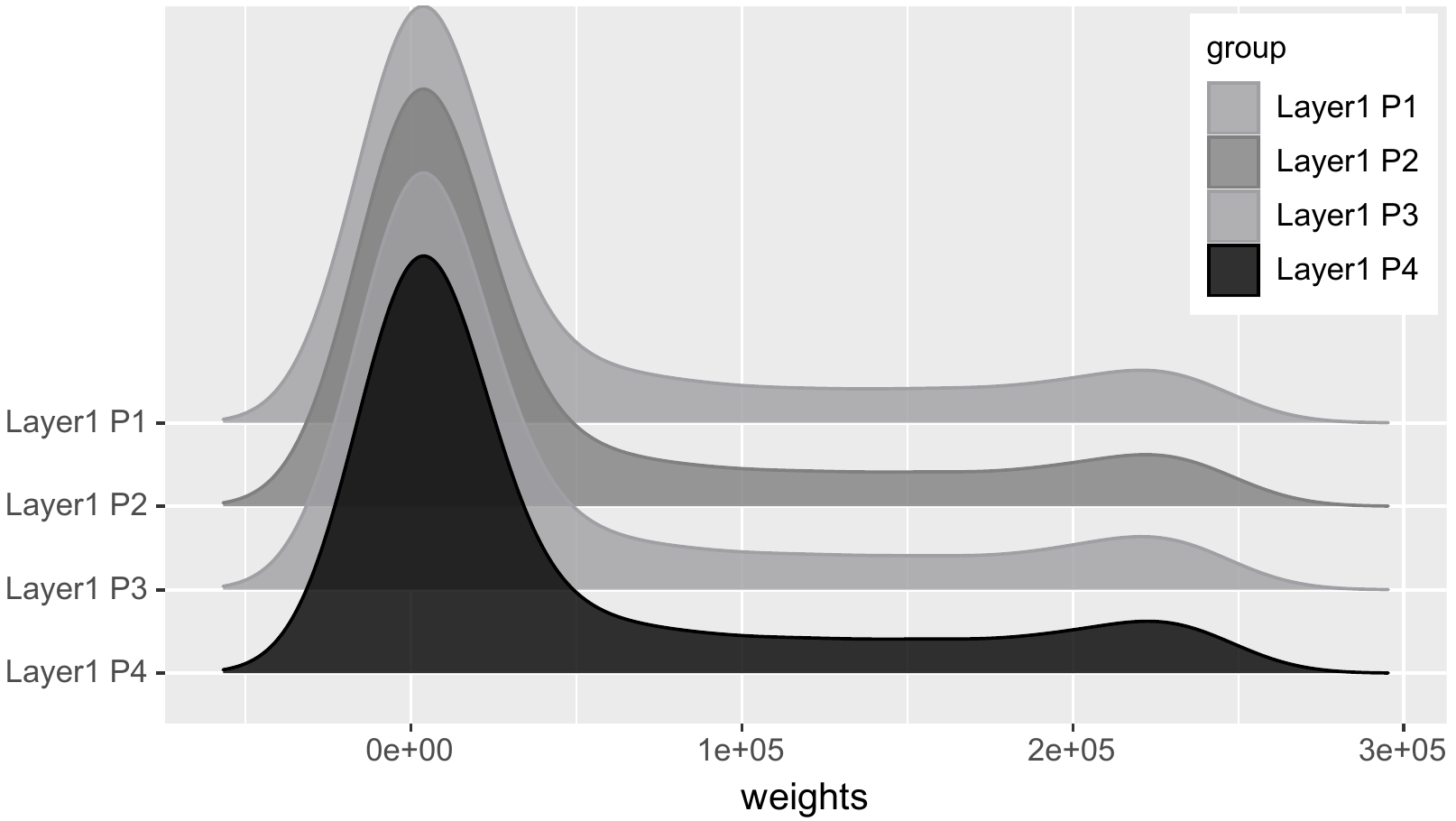}
    		\end{minipage}
		\label{fig:grid_4figs_1cap_4subcap_2}
    	}
	\subfigure[Layer1-Real]{
		\begin{minipage}[b]{0.31\textwidth}
			\includegraphics[width=1\textwidth]{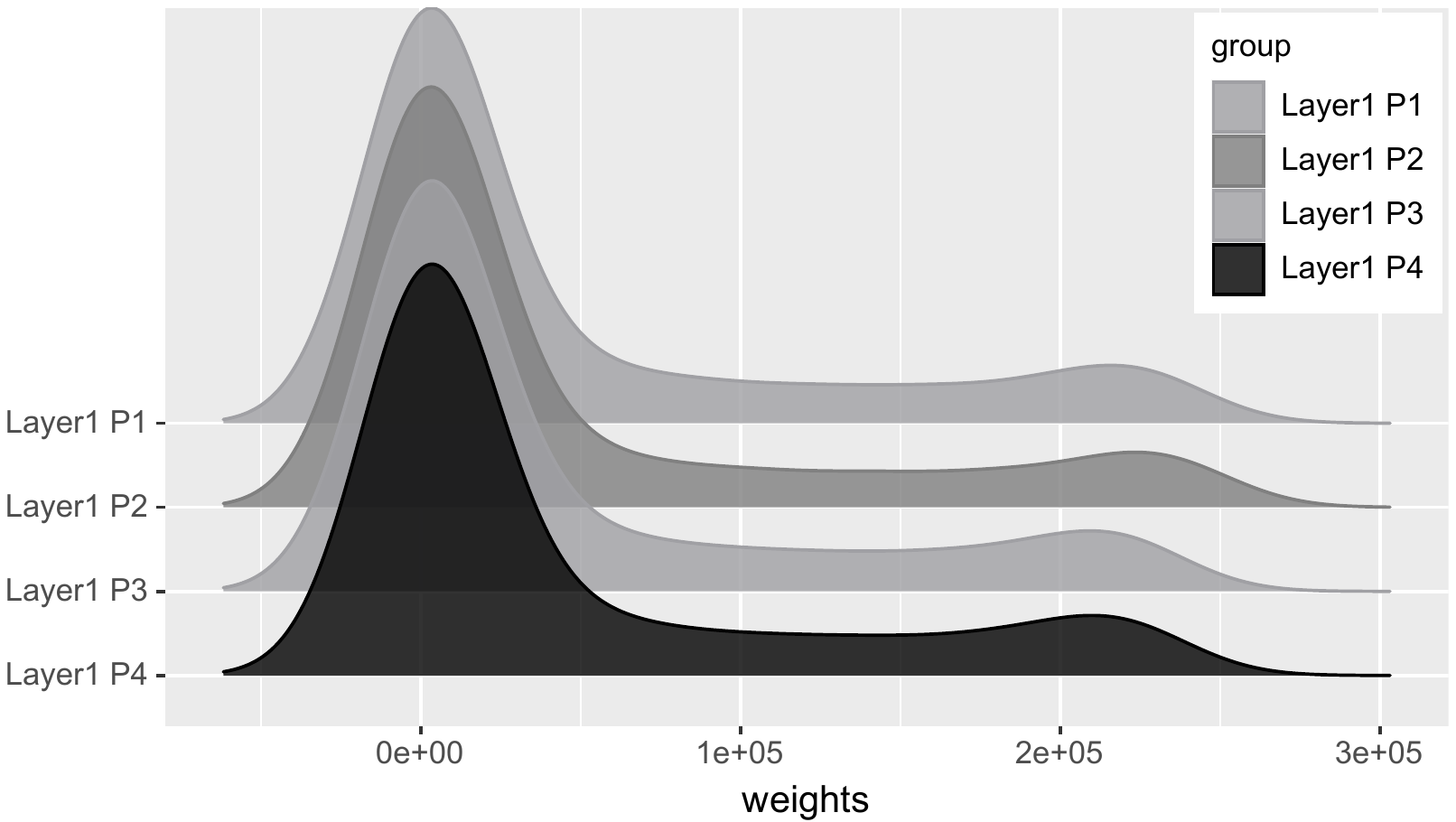} 
		\end{minipage}
		\label{fig:grid_4figs_1cap_4subcap_3}
	}
    \subfigure[Layer2-ImageNet]{
		\begin{minipage}[b]{0.31\textwidth}
			\includegraphics[width=1\textwidth]{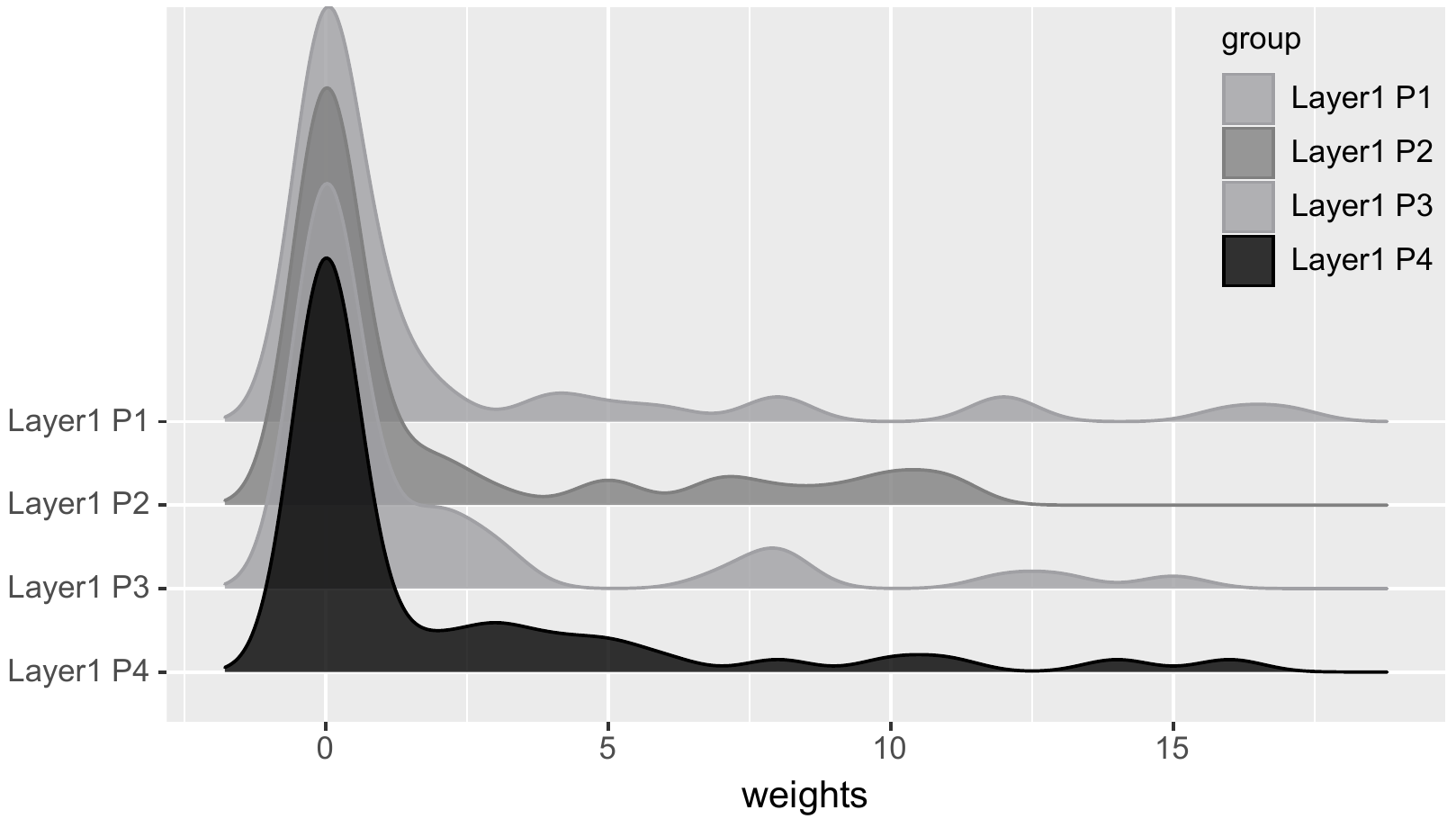} 
		\end{minipage}
		\label{fig:grid_4figs_1cap_4subcap_1}
	}
    	\subfigure[Layer2-GTAH]{
    		\begin{minipage}[b]{0.31\textwidth}
   		 	\includegraphics[width=1\textwidth]{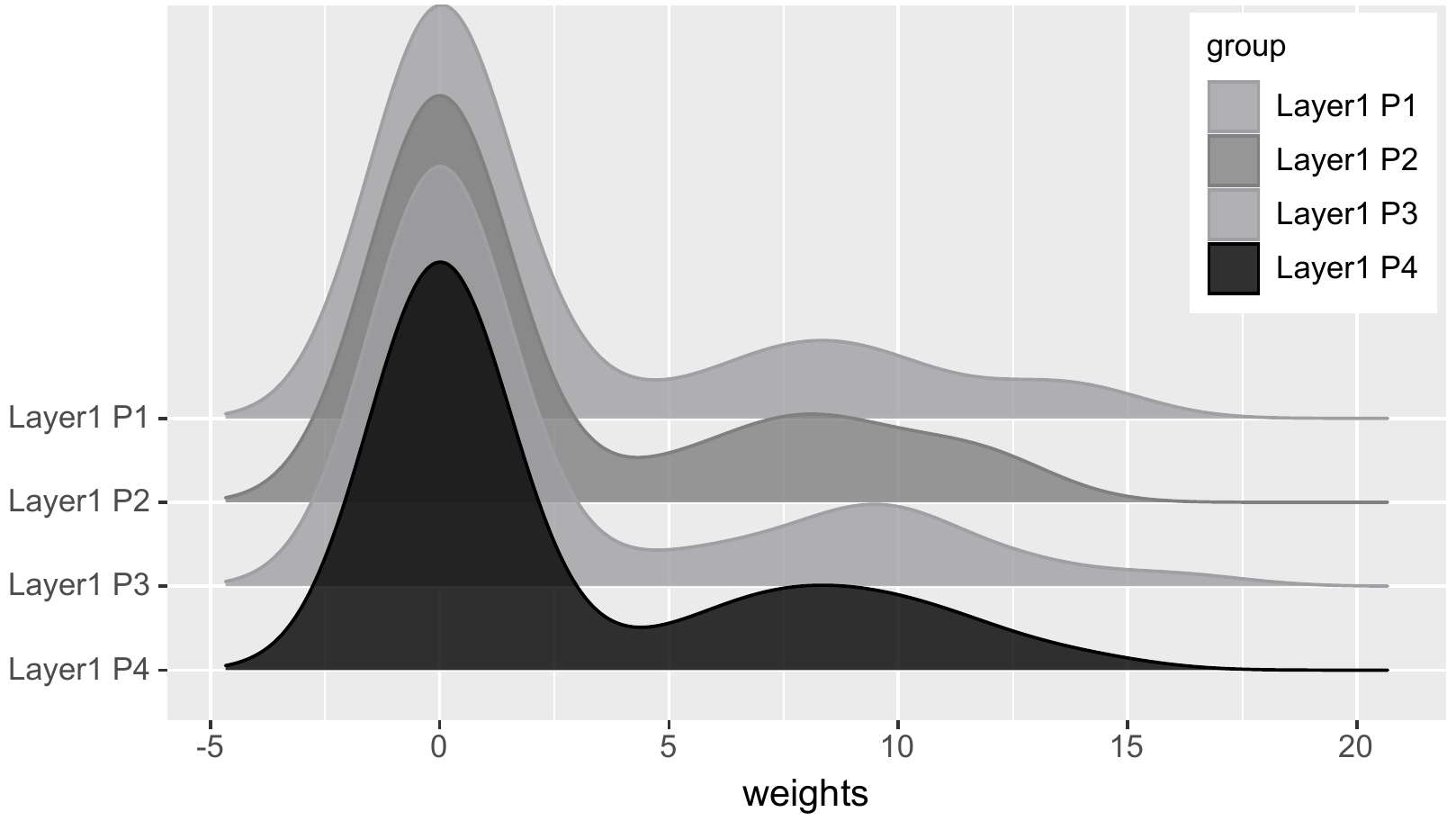}
    		\end{minipage}
		\label{fig:grid_4figs_1cap_4subcap_2}
    	}
	\subfigure[Layer2-Real]{
		\begin{minipage}[b]{0.31\textwidth}
			\includegraphics[width=1\textwidth]{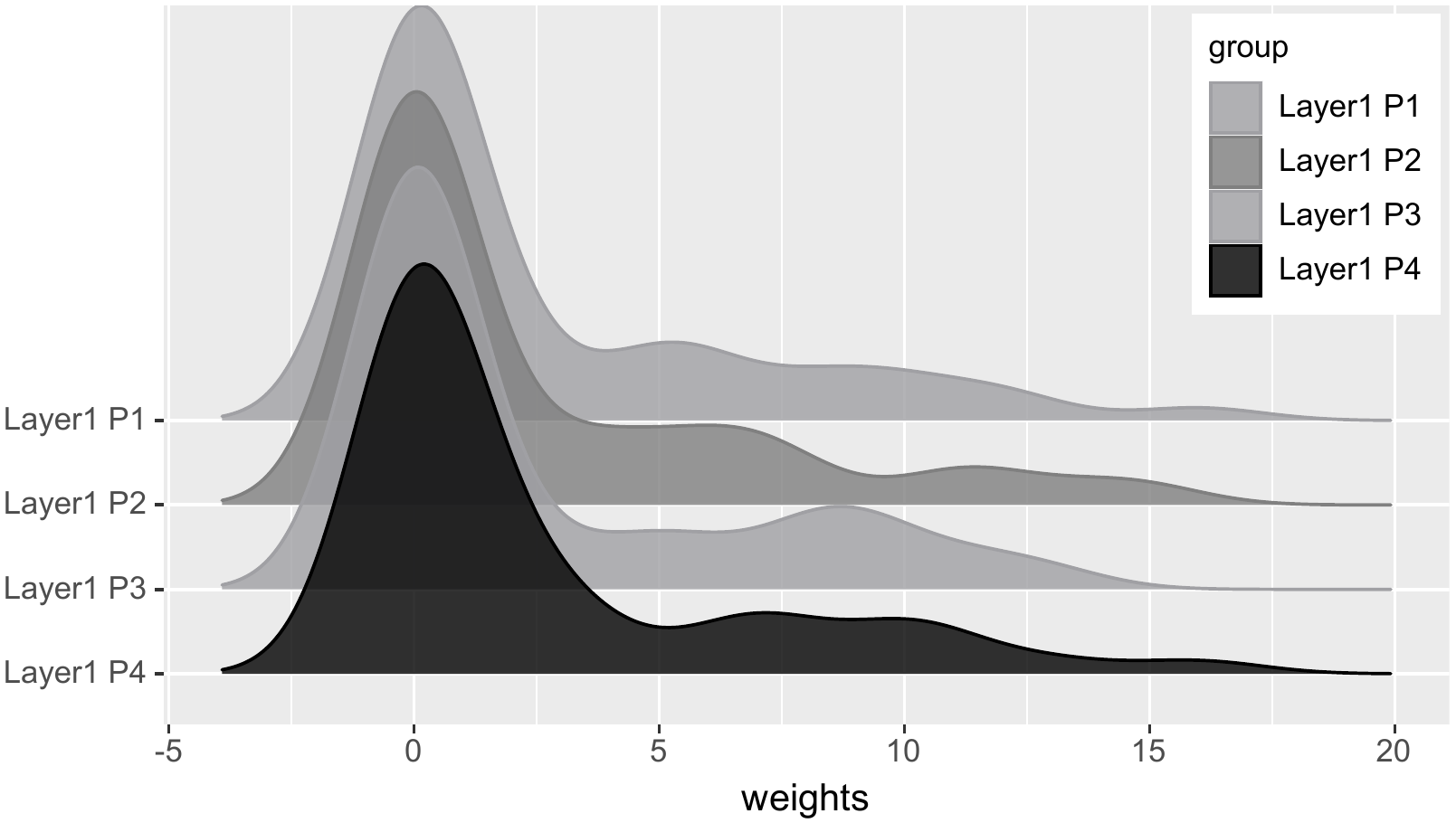} 
		\end{minipage}
		\label{fig:grid_4figs_1cap_4subcap_3}
	}
	\caption{Weight distributions of the Swin Transformer trained on ImageNet, GTAH, and the combination of real-world datasets, respectively.}
	\label{fig:parameter_distribution}
\end{figure*}
\textbf{Visualization of the Loss Tendency:} In Fig. \ref{fig:train_loss}, we can observe that for the datasets of AHN and ARG, both ImageNet and GTAH pre-trained parameters can accelerate model training, while using GTAH can result in a faster convergence rate at the early stage. Taking into consideration their test results in Table \ref{table:few_AHN} and Table \ref{table:few_ARG}, the Swin Transformer pre-trained on the GTAH dataset still outperforms that pre-trained on ImageNet. When we turn to the JAX and ATL datasets, the model using GTAH pre-trained parameters experiences a rapid decline and show a dramatic advantage to ImageNet. From the perspective of loss tendency, our proposed GTAH dataset can facilitate the training process of all the datasets, albeit to different extents.

\textbf{Visualization of the Weight Distribution:} We also visualize and compare the weight distributions of deep models trained on ImageNet, GTAH, and real-world datasets, respectively. As presented in Fig. \ref{fig:parameter_distribution}, the weight distribution of the last two layers (Layer1 and Layer2) of the SwinUper method are visualized. For both layers, parameters trained on GTAH have a more similar distribution to those trained on real-world datasets than the ImageNet pre-trained parameters. Especially for Layer1, the distribution of weights trained on GTAH is highly consistent with those trained on real datasets. For Layer2, the differences in weight distributions between ImageNet and real-world datasets become larger. This is reasonable because the shallow layers mainly extract the universal representations, while the final layer is usually responsible for the dataset-specific predictions.

\textbf{Comparison with Pre-trained Weights on Remote Sensing Data:} In Table \ref{table:pretrain_AHN}, we also compare the performance of weights pre-trained on GTAH with other remote sensing datasets. Considering that natural images in ImageNet are different from remotely-sensed images, we compare our method with SSLTransformerRS \cite{scheibenreif2022self} that are pre-trained on the Sentinel-2 dataset using self-supervised learning. As shown in Table \ref{table:pretrain_AHN}, SwinUper (SSLTransformerRS) is slightly better than SwinUper (ImageNet). Pre-training using GTAH can achieve Superior performance than others, which indicates the effectiveness of the proposed dataset.

\section{Conclusion}
\label{conclusion}
In this paper, we study the transferability of height estimation models in a cross-dataset transfer setting. To start with, a new large-scale synthetic dataset, named GTAH, for height estimation from monocular remote sensing images is constructed and released. GTAH contains highly accurate  high-resolution RGB/height image pairs captured under different imaging conditions, which can be helpful to foster research on MHE. Furthermore, we also collect and release a large-scale real-world dataset termed AHN, for the MHE task. Then, we study the transferability of deep learning models for MHE in a cross-dataset setting, which is more consistent with real-world applications. To achieve this goal, a large-scale benchmark dataset for cross-dataset transfer learning on the MHE task is constructed. Furthermore, a new experimental protocol, few-shot cross-dataset transfer, is designed to evaluate the generalizability of MHE models in a cross-dataset setting. In addition, a scale-deformable convolution module is designed to handle the severe scale variation problem. Experimental results have verified the effectiveness of the proposed new datasets and methods for the height estimation task.

\bibliographystyle{IEEEtran}
\bibliography{IEEEabrv,Reference}

% Generated by IEEEtran.bst, version: 1.12 (2007/01/11)
\begin{thebibliography}{10}
\providecommand{\url}[1]{#1}
\csname url@samestyle\endcsname
\providecommand{\newblock}{\relax}
\providecommand{\bibinfo}[2]{#2}
\providecommand{\BIBentrySTDinterwordspacing}{\spaceskip=0pt\relax}
\providecommand{\BIBentryALTinterwordstretchfactor}{4}
\providecommand{\BIBentryALTinterwordspacing}{\spaceskip=\fontdimen2\font plus
\BIBentryALTinterwordstretchfactor\fontdimen3\font minus
  \fontdimen4\font\relax}
\providecommand{\BIBforeignlanguage}[2]{{%
\expandafter\ifx\csname l@#1\endcsname\relax
\typeout{** WARNING: IEEEtran.bst: No hyphenation pattern has been}%
\typeout{** loaded for the language `#1'. Using the pattern for}%
\typeout{** the default language instead.}%
\else
\language=\csname l@#1\endcsname
\fi
#2}}
\providecommand{\BIBdecl}{\relax}
\BIBdecl

\bibitem{mou2018im2height}
L.~Mou and X.~X. Zhu, ``Im2height: Height estimation from single monocular
  imagery via fully residual convolutional-deconvolutional network,''
  \emph{arXiv preprint arXiv:1802.10249}, 2018.

\bibitem{navalgund2007remote}
R.~R. Navalgund, V.~Jayaraman, and P.~Roy, ``Remote sensing applications: an
  overview,'' \emph{current science}, pp. 1747--1766, 2007.

\bibitem{masser2001managing}
I.~Masser, ``Managing our urban future: the role of remote sensing and
  geographic information systems,'' \emph{Habitat international}, vol.~25,
  no.~4, pp. 503--512, 2001.

\bibitem{lecun1998gradient}
Y.~LeCun, L.~Bottou, Y.~Bengio, and P.~Haffner, ``Gradient-based learning
  applied to document recognition,'' \emph{Proceedings of the IEEE}, vol.~86,
  no.~11, pp. 2278--2324, 1998.

\bibitem{yosinski2014transferable}
J.~Yosinski, J.~Clune, Y.~Bengio, and H.~Lipson, ``How transferable are
  features in deep neural networks?'' \emph{arXiv preprint arXiv:1411.1792},
  2014.

\bibitem{GTAV}
Rockstar, ``Grand theft auto v,'' \url{www.rockstargames.com}, 2021.

\bibitem{saxena2005learning}
A.~Saxena, S.~H. Chung, A.~Y. Ng \emph{et~al.}, ``Learning depth from single
  monocular images,'' in \emph{NIPS}, vol.~18, 2005, pp. 1--8.

\bibitem{saxena2008make3d}
A.~Saxena, M.~Sun, and A.~Y. Ng, ``Make3d: Learning 3d scene structure from a
  single still image,'' \emph{IEEE transactions on pattern analysis and machine
  intelligence}, vol.~31, no.~5, pp. 824--840, 2008.

\bibitem{konrad20122d}
J.~Konrad, M.~Wang, and P.~Ishwar, ``2d-to-3d image conversion by learning
  depth from examples,'' in \emph{2012 IEEE Computer Society Conference on
  Computer Vision and Pattern Recognition Workshops}.\hskip 1em plus 0.5em
  minus 0.4em\relax IEEE, 2012, pp. 16--22.

\bibitem{eigen2015predicting}
D.~Eigen and R.~Fergus, ``Predicting depth, surface normals and semantic labels
  with a common multi-scale convolutional architecture,'' in \emph{Proceedings
  of the IEEE international conference on computer vision}, 2015, pp.
  2650--2658.

\bibitem{zhu2020edge}
S.~Zhu, G.~Brazil, and X.~Liu, ``The edge of depth: Explicit constraints
  between segmentation and depth,'' in \emph{Proceedings of the IEEE/CVF
  Conference on Computer Vision and Pattern Recognition}, 2020, pp.
  13\,116--13\,125.

\bibitem{jiao2018look}
J.~Jiao, Y.~Cao, Y.~Song, and R.~Lau, ``Look deeper into depth: Monocular depth
  estimation with semantic booster and attention-driven loss,'' in
  \emph{Proceedings of the European conference on computer vision (ECCV)},
  2018, pp. 53--69.

\bibitem{miangoleh2021boosting}
S.~M.~H. Miangoleh, S.~Dille, L.~Mai, S.~Paris, and Y.~Aksoy, ``Boosting
  monocular depth estimation models to high-resolution via content-adaptive
  multi-resolution merging,'' in \emph{Proceedings of the IEEE/CVF Conference
  on Computer Vision and Pattern Recognition}, 2021, pp. 9685--9694.

\bibitem{garg2016unsupervised}
R.~Garg, V.~K. Bg, G.~Carneiro, and I.~Reid, ``Unsupervised cnn for single view
  depth estimation: Geometry to the rescue,'' in \emph{European conference on
  computer vision}.\hskip 1em plus 0.5em minus 0.4em\relax Springer, 2016, pp.
  740--756.

\bibitem{godard2017unsupervised}
C.~Godard, O.~Mac~Aodha, and G.~J. Brostow, ``Unsupervised monocular depth
  estimation with left-right consistency,'' in \emph{Proceedings of the IEEE
  conference on computer vision and pattern recognition}, 2017, pp. 270--279.

\bibitem{pilzer2019refine}
A.~Pilzer, S.~Lathuiliere, N.~Sebe, and E.~Ricci, ``Refine and distill:
  Exploiting cycle-inconsistency and knowledge distillation for unsupervised
  monocular depth estimation,'' in \emph{Proceedings of the IEEE/CVF Conference
  on Computer Vision and Pattern Recognition}, 2019, pp. 9768--9777.

\bibitem{peng2021excavating}
R.~Peng, R.~Wang, Y.~Lai, L.~Tang, and Y.~Cai, ``Excavating the potential
  capacity of self-supervised monocular depth estimation,'' in
  \emph{Proceedings of the IEEE/CVF International Conference on Computer
  Vision}, 2021, pp. 15\,560--15\,569.

\bibitem{jung2021fine}
H.~Jung, E.~Park, and S.~Yoo, ``Fine-grained semantics-aware representation
  enhancement for self-supervised monocular depth estimation,'' in
  \emph{Proceedings of the IEEE/CVF International Conference on Computer
  Vision}, 2021, pp. 12\,642--12\,652.

\bibitem{jiao2021effiscene}
Y.~Jiao, T.~D. Tran, and G.~Shi, ``Effiscene: Efficient per-pixel rigidity
  inference for unsupervised joint learning of optical flow, depth, camera pose
  and motion segmentation,'' in \emph{Proceedings of the IEEE/CVF Conference on
  Computer Vision and Pattern Recognition}, 2021, pp. 5538--5547.

\bibitem{tosi2020distilled}
F.~Tosi, F.~Aleotti, P.~Z. Ramirez, M.~Poggi, S.~Salti, L.~D. Stefano, and
  S.~Mattoccia, ``Distilled semantics for comprehensive scene understanding
  from videos,'' in \emph{Proceedings of the IEEE/CVF Conference on Computer
  Vision and Pattern Recognition}, 2020, pp. 4654--4665.

\bibitem{godard2019digging}
C.~Godard, O.~Mac~Aodha, M.~Firman, and G.~J. Brostow, ``Digging into
  self-supervised monocular depth estimation,'' in \emph{Proceedings of the
  IEEE/CVF International Conference on Computer Vision}, 2019, pp. 3828--3838.

\bibitem{shu2020feature}
C.~Shu, K.~Yu, Z.~Duan, and K.~Yang, ``Feature-metric loss for self-supervised
  learning of depth and egomotion,'' in \emph{European Conference on Computer
  Vision}.\hskip 1em plus 0.5em minus 0.4em\relax Springer, 2020, pp. 572--588.

\bibitem{guizilini20203d}
V.~Guizilini, R.~Ambrus, S.~Pillai, A.~Raventos, and A.~Gaidon, ``3d packing
  for self-supervised monocular depth estimation,'' in \emph{Proceedings of the
  IEEE/CVF Conference on Computer Vision and Pattern Recognition}, 2020, pp.
  2485--2494.

\bibitem{lyu2021hr}
X.~Lyu, L.~Liu, M.~Wang, X.~Kong, L.~Liu, Y.~Liu, X.~Chen, and Y.~Yuan,
  ``Hr-depth: High resolution self-supervised monocular depth estimation,'' in
  \emph{Proceedings of the AAAI Conference on Artificial Intelligence},
  vol.~35, no.~3, 2021, pp. 2294--2301.

\bibitem{guizilini2019semantically}
V.~Guizilini, R.~Hou, J.~Li, R.~Ambrus, and A.~Gaidon, ``Semantically-guided
  representation learning for self-supervised monocular depth,'' in
  \emph{International Conference on Learning Representations}, 2019.

\bibitem{liu2015learning}
F.~Liu, C.~Shen, G.~Lin, and I.~Reid, ``Learning depth from single monocular
  images using deep convolutional neural fields,'' \emph{IEEE transactions on
  pattern analysis and machine intelligence}, vol.~38, no.~10, pp. 2024--2039,
  2015.

\bibitem{laina2016deeper}
I.~Laina, C.~Rupprecht, V.~Belagiannis, F.~Tombari, and N.~Navab, ``Deeper
  depth prediction with fully convolutional residual networks,'' in \emph{2016
  Fourth international conference on 3D vision (3DV)}.\hskip 1em plus 0.5em
  minus 0.4em\relax IEEE, 2016, pp. 239--248.

\bibitem{chen2019towards}
P.-Y. Chen, A.~H. Liu, Y.-C. Liu, and Y.-C.~F. Wang, ``Towards scene
  understanding: Unsupervised monocular depth estimation with semantic-aware
  representation,'' in \emph{Proceedings of the IEEE/CVF Conference on Computer
  Vision and Pattern Recognition}, 2019, pp. 2624--2632.

\bibitem{wang2020sdc}
L.~Wang, J.~Zhang, O.~Wang, Z.~Lin, and H.~Lu, ``Sdc-depth: Semantic
  divide-and-conquer network for monocular depth estimation,'' in
  \emph{Proceedings of the IEEE/CVF Conference on Computer Vision and Pattern
  Recognition}, 2020, pp. 541--550.

\bibitem{yin2019enforcing}
W.~Yin, Y.~Liu, C.~Shen, and Y.~Yan, ``Enforcing geometric constraints of
  virtual normal for depth prediction,'' in \emph{Proceedings of the IEEE/CVF
  International Conference on Computer Vision}, 2019, pp. 5684--5693.

\bibitem{lee2020multi}
J.-H. Lee and C.-S. Kim, ``Multi-loss rebalancing algorithm for monocular depth
  estimation,'' in \emph{Computer Vision--ECCV 2020: 16th European Conference,
  Glasgow, UK, August 23--28, 2020, Proceedings, Part XVII 16}.\hskip 1em plus
  0.5em minus 0.4em\relax Springer, 2020, pp. 785--801.

\bibitem{ranftl2019towards}
R.~Ranftl, K.~Lasinger, D.~Hafner, K.~Schindler, and V.~Koltun, ``Towards
  robust monocular depth estimation: Mixing datasets for zero-shot
  cross-dataset transfer,'' \emph{IEEE transactions on pattern analysis and
  machine intelligence, doi: 10.1109/TPAMI.2020.3019967.}, 2019.

\bibitem{ranftl2021vision}
R.~Ranftl, A.~Bochkovskiy, and V.~Koltun, ``Vision transformers for dense
  prediction,'' in \emph{Proceedings of the IEEE/CVF International Conference
  on Computer Vision}, 2021, pp. 12\,179--12\,188.

\bibitem{li2021revisiting}
Z.~Li, X.~Liu, N.~Drenkow, A.~Ding, F.~X. Creighton, R.~H. Taylor, and
  M.~Unberath, ``Revisiting stereo depth estimation from a sequence-to-sequence
  perspective with transformers,'' in \emph{Proceedings of the IEEE/CVF
  International Conference on Computer Vision}, 2021, pp. 6197--6206.

\bibitem{srivastava2017joint}
S.~Srivastava, M.~Volpi, and D.~Tuia, ``Joint height estimation and semantic
  labeling of monocular aerial images with cnns,'' in \emph{2017 IEEE
  International Geoscience and Remote Sensing Symposium (IGARSS)}.\hskip 1em
  plus 0.5em minus 0.4em\relax IEEE, 2017, pp. 5173--5176.

\bibitem{ghamisi2018img2dsm}
P.~Ghamisi and N.~Yokoya, ``Img2dsm: Height simulation from single imagery
  using conditional generative adversarial net,'' \emph{IEEE Geoscience and
  Remote Sensing Letters}, vol.~15, no.~5, pp. 794--798, 2018.

\bibitem{paoletti2020u}
M.~Paoletti, J.~Haut, P.~Ghamisi, N.~Yokoya, J.~Plaza, and A.~Plaza,
  ``U-img2dsm: Unpaired simulation of digital surface models with generative
  adversarial networks,'' \emph{IEEE Geoscience and Remote Sensing Letters},
  2020.

\bibitem{amini2019cnn}
H.~Amini~Amirkolaee and H.~Arefi, ``Cnn-based estimation of pre-and
  post-earthquake height models from single optical images for identification
  of collapsed buildings,'' \emph{Remote Sensing Letters}, vol.~10, no.~7, pp.
  679--688, 2019.

\bibitem{amirkolaee2019height}
H.~A. Amirkolaee and H.~Arefi, ``Height estimation from single aerial images
  using a deep convolutional encoder-decoder network,'' \emph{ISPRS journal of
  photogrammetry and remote sensing}, vol. 149, pp. 50--66, 2019.

\bibitem{liu2021hecr}
W.~Liu, W.~Zhang, X.~Sun, Z.~Guo, and K.~Fu, ``Hecr-net: Height-embedding
  context reassembly network for semantic segmentation in aerial images,''
  \emph{IEEE Journal of Selected Topics in Applied Earth Observations and
  Remote Sensing}, vol.~14, pp. 9117--9131, 2021.

\bibitem{christie2020learning}
G.~Christie, R.~R. R.~M. Abujder, K.~Foster, S.~Hagstrom, G.~D. Hager, and
  M.~Z. Brown, ``Learning geocentric object pose in oblique monocular images,''
  in \emph{Proceedings of the IEEE/CVF Conference on Computer Vision and
  Pattern Recognition}, 2020, pp. 14\,512--14\,520.

\bibitem{mahmud2020boundary}
J.~Mahmud, T.~Price, A.~Bapat, and J.-M. Frahm, ``Boundary-aware 3d building
  reconstruction from a single overhead image,'' in \emph{Proceedings of the
  IEEE/CVF Conference on Computer Vision and Pattern Recognition}, 2020, pp.
  441--451.

\bibitem{madhuanand2021self}
L.~Madhuanand, F.~Nex, and M.~Y. Yang, ``Self-supervised monocular depth
  estimation from oblique uav videos,'' \emph{ISPRS Journal of Photogrammetry
  and Remote Sensing}, vol. 176, pp. 1--14, 2021.

\bibitem{zheng2019pop}
Z.~Zheng, Y.~Zhong, and J.~Wang, ``Pop-net: Encoder-dual decoder for semantic
  segmentation and single-view height estimation,'' in \emph{IGARSS 2019-2019
  IEEE International Geoscience and Remote Sensing Symposium}.\hskip 1em plus
  0.5em minus 0.4em\relax IEEE, 2019, pp. 4963--4966.

\bibitem{liu2020im2elevation}
C.-J. Liu, V.~A. Krylov, P.~Kane, G.~Kavanagh, and R.~Dahyot, ``Im2elevation:
  Building height estimation from single-view aerial imagery,'' \emph{Remote
  Sensing}, vol.~12, no.~17, p. 2719, 2020.

\bibitem{li2020height}
X.~Li, M.~Wang, and Y.~Fang, ``Height estimation from single aerial images
  using a deep ordinal regression network,'' \emph{IEEE Geoscience and Remote
  Sensing Letters}, 2020.

\bibitem{dijk2019neural}
T.~v. Dijk and G.~d. Croon, ``How do neural networks see depth in single
  images?'' in \emph{Proceedings of the IEEE/CVF International Conference on
  Computer Vision}, 2019, pp. 2183--2191.

\bibitem{christie2021geocentricpose}
G.~Christie, K.~H. Foster, S.~Hagstrom, G.~D. Hager, and M.~Z. Brown, ``Single
  view geocentric pose in the wild,'' in \emph{CVPRW}, 2021.

\bibitem{bosch2019semantic}
M.~Bosch, K.~Foster, G.~Christie, S.~Wang, G.~D. Hager, and M.~Brown,
  ``Semantic stereo for incidental satellite images,'' in \emph{2019 IEEE
  Winter Conference on Applications of Computer Vision (WACV)}.\hskip 1em plus
  0.5em minus 0.4em\relax IEEE, 2019, pp. 1524--1532.

\bibitem{kunwar2020large}
S.~Kunwar, H.~Chen, M.~Lin, H.~Zhang, P.~D’Angelo, D.~Cerra, S.~M. Azimi,
  M.~Brown, G.~Hager, N.~Yokoya \emph{et~al.}, ``Large-scale semantic 3-d
  reconstruction: Outcome of the 2019 ieee grss data fusion contest—part a,''
  \emph{IEEE Journal of Selected Topics in Applied Earth Observations and
  Remote Sensing}, vol.~14, pp. 922--935, 2020.

\bibitem{dai2017deformable}
J.~Dai, H.~Qi, Y.~Xiong, Y.~Li, G.~Zhang, H.~Hu, and Y.~Wei, ``Deformable
  convolutional networks,'' in \emph{Proceedings of the IEEE international
  conference on computer vision}, 2017, pp. 764--773.

\bibitem{Eigen2014Depth}
D.~Eigen, C.~Puhrsch, and R.~Fergus, ``Depth map prediction from a single image
  using a multi-scale deep network,'' in \emph{International Conference on
  Neural Information Processing Systems}, 2014, pp. 2366--2374.

\bibitem{chen2016single}
W.~Chen, Z.~Fu, D.~Yang, and J.~Deng, ``Single-image depth perception in the
  wild,'' \emph{Advances in neural information processing systems}, vol.~29,
  pp. 730--738, 2016.

\bibitem{wang2019web}
C.~Wang, S.~Lucey, F.~Perazzi, and O.~Wang, ``Web stereo video supervision for
  depth prediction from dynamic scenes,'' in \emph{2019 International
  Conference on 3D Vision (3DV)}.\hskip 1em plus 0.5em minus 0.4em\relax IEEE,
  2019, pp. 348--357.

\bibitem{christie2020geocentricpose}
G.~Christie, R.~Munoz, K.~H. Foster, S.~T. Hagstrom, G.~D. Hager, and M.~Z.
  Brown, ``Learning geocentric object pose in oblique monocular images,'' in
  \emph{CVPR}, 2020.

\bibitem{liu2021swin}
Z.~Liu, Y.~Lin, Y.~Cao, H.~Hu, Y.~Wei, Z.~Zhang, S.~Lin, and B.~Guo, ``Swin
  transformer: Hierarchical vision transformer using shifted windows,''
  \emph{arXiv preprint arXiv:2103.14030}, 2021.

\bibitem{kingma2014adam}
D.~P. Kingma and J.~Ba, ``Adam: A method for stochastic optimization,''
  \emph{arXiv preprint arXiv:1412.6980}, 2014.

\bibitem{xiao2018unified}
T.~Xiao, Y.~Liu, B.~Zhou, Y.~Jiang, and J.~Sun, ``Unified perceptual parsing
  for scene understanding,'' in \emph{Proceedings of the European Conference on
  Computer Vision (ECCV)}, 2018, pp. 418--434.

\bibitem{loshchilov2018fixing}
I.~Loshchilov and F.~Hutter, ``Fixing weight decay regularization in adam,''
  2018.

\bibitem{bhat2021adabins}
S.~F. Bhat, I.~Alhashim, and P.~Wonka, ``Adabins: Depth estimation using
  adaptive bins,'' in \emph{Proceedings of the IEEE/CVF Conference on Computer
  Vision and Pattern Recognition}, 2021, pp. 4009--4018.

\bibitem{ronneberger2015u}
O.~Ronneberger, P.~Fischer, and T.~Brox, ``U-net: Convolutional networks for
  biomedical image segmentation,'' in \emph{International Conference on Medical
  image computing and computer-assisted intervention}.\hskip 1em plus 0.5em
  minus 0.4em\relax Springer, 2015, pp. 234--241.

\bibitem{scheibenreif2022self}
L.~Scheibenreif, J.~Hanna, M.~Mommert, and D.~Borth, ``Self-supervised vision
  transformers for land-cover segmentation and classification,'' in
  \emph{Proceedings of the IEEE/CVF Conference on Computer Vision and Pattern
  Recognition}, 2022, pp. 1422--1431.

\end{thebibliography}
\begin{IEEEbiography}[{\includegraphics[width=1in,height=1.25in,clip,keepaspectratio]{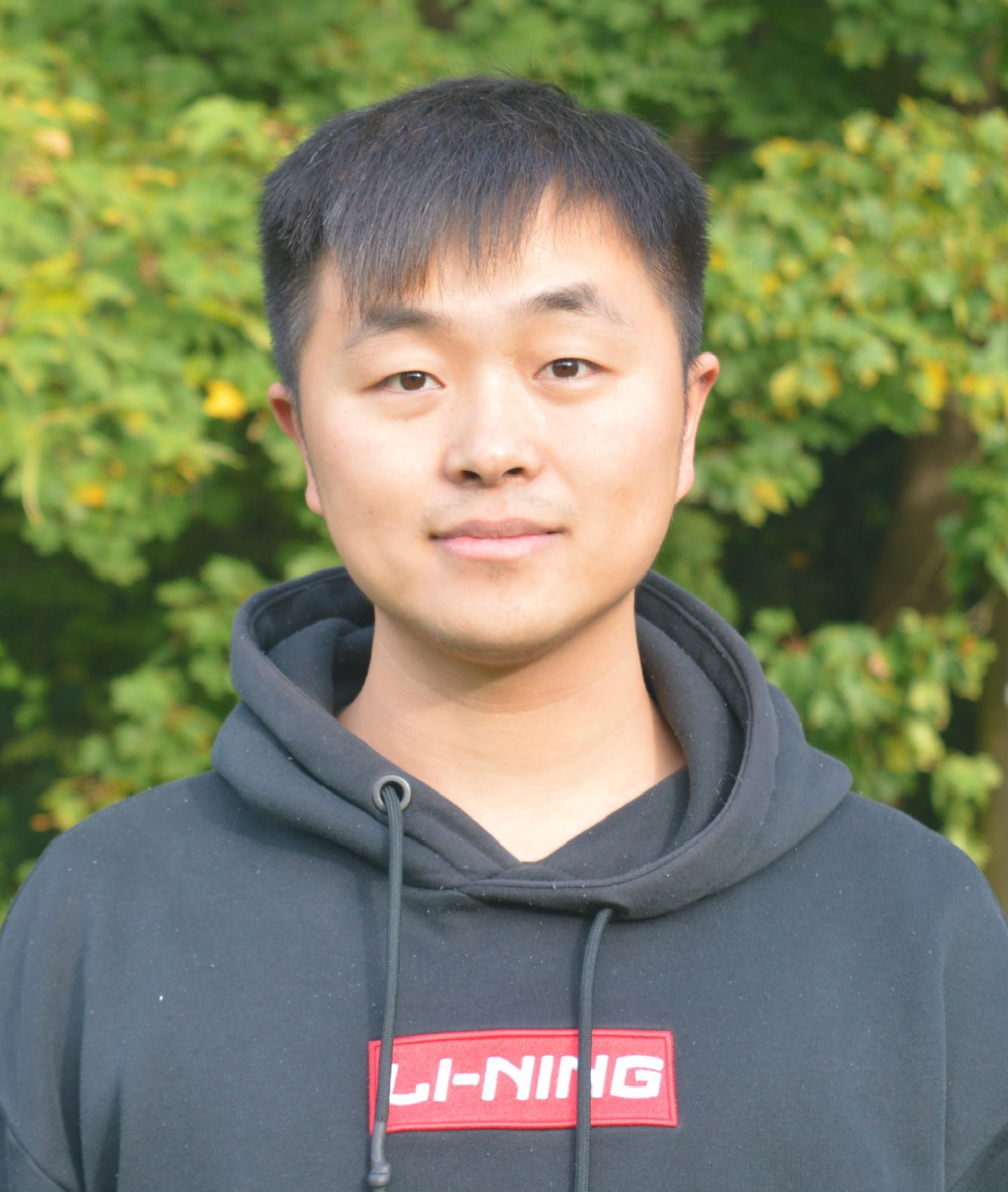}}]{Zhitong Xiong}
(Member, IEEE) received the Ph.D. degree in computer science and technology from Northwestern Polytechnical University, Xi’an, China, in 2021. He is currently a research scientist and leads the ML4Earth working group with the Data Science in Earth Observation, Technical University of Munich (TUM), Germany. His research interests include computer vision, machine learning, Earth observation, and Earth system modeling.
\end{IEEEbiography}
\vspace{-12 mm}
\begin{IEEEbiography}[{\includegraphics[width=1in,height=1.25in,clip,keepaspectratio]{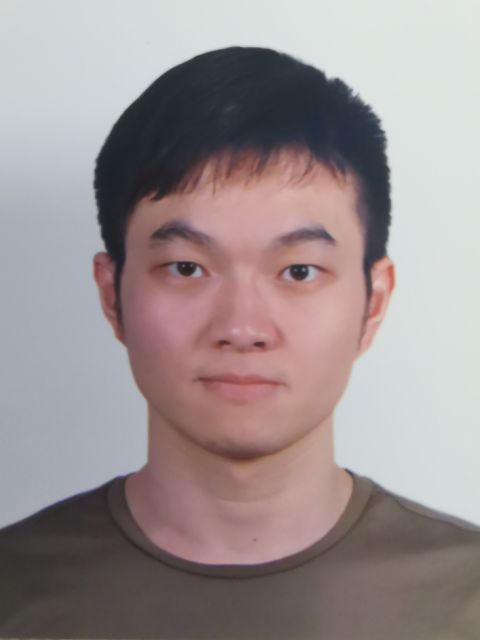}}]{Wei Huang}
received the B.E. and M.E. degree
in Northwestern Polytechnical University, Xi’an,
China, in 2018 and 2021 respectively. He is currently
working as a visiting student at Data Science in
Earth Observation, Technical University of Munich, Munich, 
Germany. His research interests include computer
vision and remote sensing.
\end{IEEEbiography}
\vspace{-10 mm}
\begin{IEEEbiography}[{\includegraphics[width=1in,height=1.25in,clip,keepaspectratio]{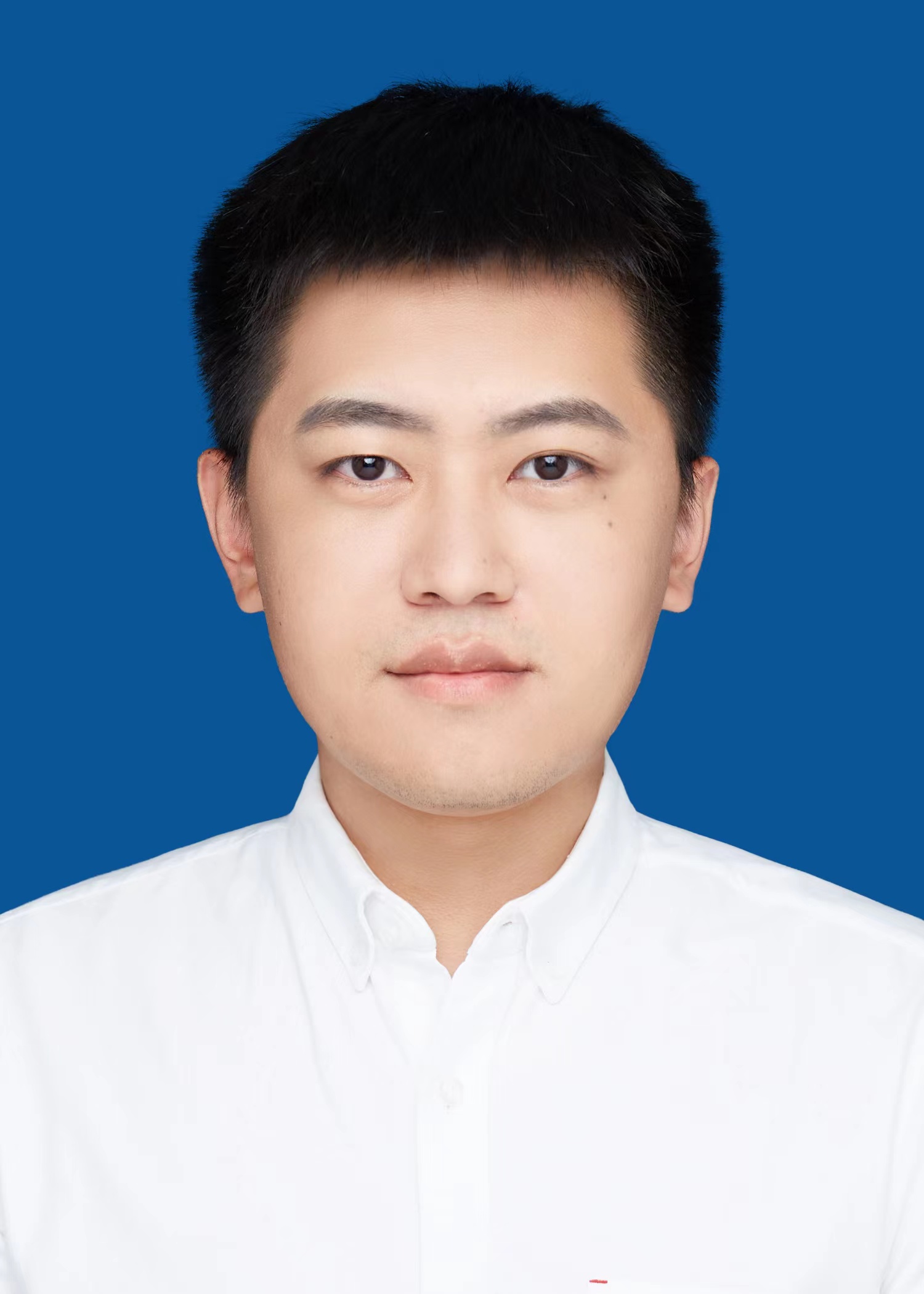}}]{Jingtao Hu}
received the M.E. degree in control engineering from the Lanzhou University of Technology, Lanzhou. He is currently pursuing the Ph.D. degree with the School of Computer Science and the School of Artificial Intelligence, Optics and Electronics (iOPEN), Northwestern Polytechnical University, Xi'an, China. His research interests include remote sensing, computer vison and machine learning.
\end{IEEEbiography}
\vspace{-10 mm}
\begin{IEEEbiography}[{\includegraphics[width=1in,height=1.25in,clip,keepaspectratio]{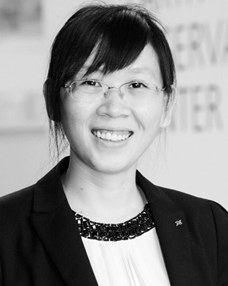}}]{Xiaoxiang Zhu}.
(S’10-M’12-SM’14-F’20) received the M.Sc., Dr.Ing., and Habilitation degrees in signal
processing from the Technical University of Munich (TUM), Munich, Germany, in 2008, 2011, and 2013,
respectively. She is currently the Professor for Data Science in Earth Observation at Technical University of Munich (TUM) and the Head of the Department ``EO Data Science'' at the Remote Sensing Technology Institute, German Aerospace Center (DLR). Since 2019, Zhu is a co-coordinator of the Munich Data Science Research School (www.mu-ds.de). Since 2019 She also heads the Helmholtz Artificial Intelligence -- Research Field ``Aeronautics, Space and Transport". Since May 2020, she is the director of the international future AI lab "AI4EO -- Artificial Intelligence for Earth Observation: Reasoning, Uncertainties, Ethics and Beyond", Munich, Germany. Since October 2020, she also serves as a co-director of the Munich Data Science Institute (MDSI), TUM. Prof. Zhu was a guest scientist or visiting professor at the Italian National Research Council (CNR-IREA), Naples, Italy, Fudan University, Shanghai, China, the University  of Tokyo, Tokyo, Japan and University of California, Los Angeles, United States in 2009, 2014, 2015 and 2016, respectively. She is currently a visiting AI professor at ESA's Phi-lab. Her main research interests are remote sensing and Earth observation, signal processing, machine learning and data science, with a special application focus on global urban mapping. 
\end{IEEEbiography}

\ifCLASSOPTIONcaptionsoff
  \newpage
\fi

\end{document}